\pdfoutput=1

\documentclass[11pt]{article}

\usepackage{ACL2023}
\usepackage{booktabs}
\usepackage{float}
\usepackage{longtable}
\usepackage{times}
\usepackage{amsthm}
\usepackage{amssymb}
\usepackage{tabularx}
\usepackage{colortbl}
\usepackage{array}
\usepackage{enumitem}
\usepackage{latexsym}
\usepackage{multirow}
\usepackage[T1]{fontenc}

\usepackage{amsmath}
\usepackage{algorithm2e}

\usepackage[utf8]{inputenc}

\usepackage{microtype}

\usepackage{inconsolata}
\usepackage{tabularx}
\usepackage{amsmath}
\usepackage{mathtools}
\usepackage{amssymb}
\usepackage{amsfonts}
\usepackage{url}
\newcolumntype{Y}{>{\centering\arraybackslash}X}
\newtheorem{definition}{Definition}[section]

\title{Controlled LLM-based Reasoning for Clinical Trial Retrieval}

\author{Mael Jullien$^1$ \and Alex Bogatu$^2$ \and Harriet Unsworth$^2$ \and Andr\'e Freitas$^{1,2,3}$\\
$^{1}$ Department of Computer Science, University of Manchester, United Kingdom\\ 
$^{2}$ Cancer Biomarker Centre, CRUK Manchester Institute, United Kingdom\\ 
$^{3}$ Idiap Research Institute, Switzerland}

\begin{document}
\theoremstyle{definition}
\maketitle
\begin{abstract}

Matching patients to clinical trials demands a systematic and reasoned interpretation of documents which require significant expert-level background knowledge, over a complex set of well-defined eligibility criteria. Moreover, this interpretation process needs to operate at scale, over vast knowledge bases of trials. In this paper, we propose a scalable method that extends the capabilities of LLMs in the direction of systematizing the reasoning over sets of medical eligibility criteria, evaluating it in the context of real-world cases. The proposed method overlays a Set-guided reasoning method for LLMs. The proposed framework is evaluated on TREC 2022 Clinical Trials, achieving results superior to the state-of-the-art: NDCG@10 of 0.693 and Precision@10 of 0.73.
\end{abstract}

\section{Introduction} \label{sec:intro}

Patient recruitment remains a major barrier for clinical trials, despite significant efforts invested in tackling this challenge in the last decade\footnote{National Institute of Health (NIH) notes that approximately 80\% of clinical trials do not meet their recruitment targets, often prolonging or derailing medical research}. To address this, the employment of Natural Language Processing (NLP) methods, most notably the use of Large Language Models (LLMs) \cite{yuan2023llm}, has shown promising results, motivating the setup of several pilot studies to explore the use of such models for clinical trial records (CTR) retrieval \cite{Roberts2022OverviewOT, yuan2023llm}. 

In practice however, most of the proposed methods still have limited performance with regard to precision and recall, which carries significant ethical implications for clinical applications. Additionally, these methods lack interpretability and control, where the initial hypotheses of being able to analyze a patient’s medical history or to understand a trial's eligibility criteria is yet to be confirmed. 

This paper introduces a novel method for systematic reasoning of trial retrieval and re-ranking that leverages LLMs prompting to initially transform unstructured patient notes and CTRs into attribute sets, thus facilitating the precise mapping, interpretability and scalable retrieval of CTRs that are directly related to a patient’s specific medical conditions. The resulting sets of attributes enable a set-guided reasoning process that can deliver a first-stage trial retrieval, expanding the traditional similarity-based approaches \cite{Kusa2023EffectiveMO, peikos2023utilizing} with hierarchical relationships conveyed by the use of domain-specific knowledge. This overlay of a structured, set-theoretical perspective to LLMs allows for a step-wise and controlled LLM-based inference method for candidate trial interpretation and matching, which allows for a systematic interpretation of both granular (i.e., individual-level) and broad (i.e., entire CTR-level) eligibility criteria. Finally, we perform an analytical exploration of a space of ranking functions, each designed to aggregate the previous results in an explainable and controllable manner, integrated with an LLM-based delivered deontic reasoning. Broadly, the contributions of this paper include: 
\begin{itemize}[leftmargin=*,noitemsep,nolistsep]
    \item A set-guided reasoning framework for LLM-based retrieval that ensures generation grounded in ontological knowledge.
    \item A deontic-style reasoning over clinical trial eligibility that explores several ranking functions and leads to interpretable selection of CTRs.
    \item An extensive evaluation and analysis of the proposed modeling interventions, individually and against the state-of-the-art, based on TREC 2022.
\end{itemize}

\begin{figure*}[t]
\centering
\includegraphics[width=.8\textwidth]{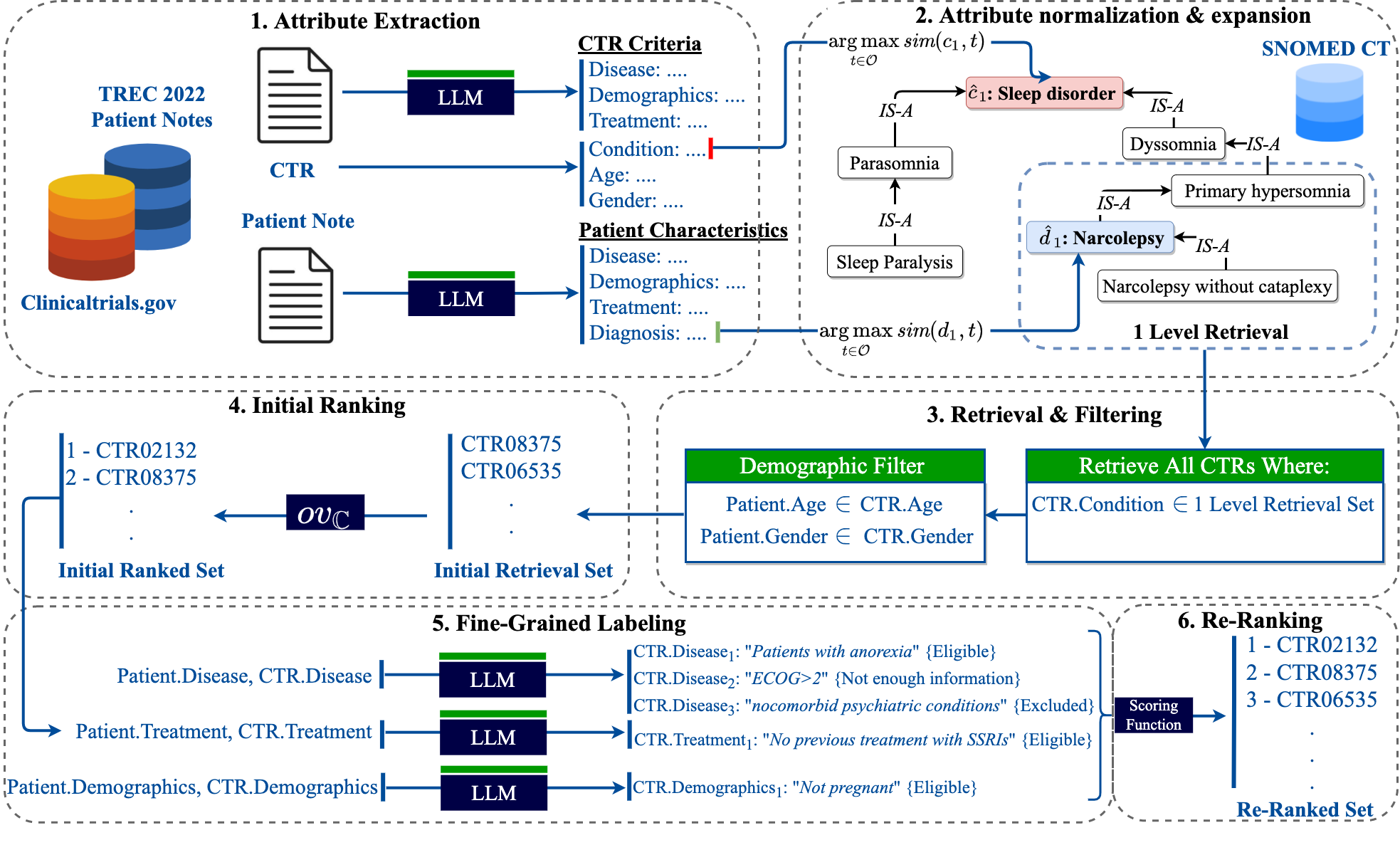}

\caption{End-to-end depiction of the set-reasoning guided patient-trial matching.}
\label{fig:ontogenr}
\end{figure*}

\section{Ontology-grounded LLM-based CT retrieval}


The complexity of CT retrieval can be summarized by the following factors: (a) the need to interpret complex sets of multiple eligibility criteria statements (both inclusion and exclusion), (b) the need to interpret domain-specific concepts and definitions, (c) the requirement to cross the semantic and abstraction gap between patients and CT descriptions and (d) the need to perform this interpretation at scale. 

The spirit of this interpretation is intensional (in a set-theoretical sense) and granular in nature: it involves the reasoning over concepts, definitions and criteria statements, which will deliver a result set of matching treatments for a given patient. Descriptions of patients and CTRs are largely variable in their terminological expression and abstraction level. Moreover, the task is of a retrieval nature, where knowledge bases of CTRs are of the order of hundreds of thousands documents. Additionally, the task is recall-dominant, where the priority is on the retrieval of all options available to that patient. This defines a natural tension between the required semantic granularity, terminological specialization and systematic control and the lack of granularity of text embedding and keyword-based methods, which are prevalent in retrieval and retrieval augmented generation (RAG) methods \cite{Kusa2023EffectiveMO, peikos2023utilizing, Sin2022JBNUAT}.

In this work we address this tension by overlaying an intensional reasoning layer within an LLM-supported CT retrieval model. This consists on the composition of three major modeling strategies, illustrated in Figure \ref{fig:ontogenr}: (i) structuring and typing the unstructured content (patient and clinical trial description), where patient data and CTR information are parsed into sets of predicates and axioms, with the aim of addressing challenge (a) above; (ii) the integration of specialized ontologies within the retrieval process, which enriches the inference process with specialized concepts, with the aim of addressing (b) above; and (iii) with the support of its taxonomic structure and synonymic sets can support addressing the terminological and abstraction gap (c), while keeping scalability. This is complemented by the use of targeted criteria filters. The previous steps aim at delivering a scalable high-recall result set, which allows for a guided prompt analysis of the eligibility criteria sets using deontic prompts.

Overall, this methodology underscores the alignment between structuring patient and CT descriptions, leveraging ontological relationships for precise retrieval at scale, intensional-level reasoning and the integration of deontic prompt-based reasoning for interpretable clinical trial selection.

\section{Domain-specific attribute extraction}

The first step in our proposed retrieval framework consists of structuring patient note and CTR data based on a collection of predefined attributes (e.g., disease, demographics, treatment, etc.), with values construed as sets that collectively describe the characteristics of the input data. This process is performed via LLM-prompting and domain-specific ontology mapping.

\subsection{Patient notes modeling}

In practice, patient notes predominantly consist of succinct free-text case descriptions, with some structured lists of test results values. Specifically, a patient note can be represented as a collection of attributes: \textit{Age} ($\mathbb{A}$), \textit{Gender} ($\mathbb{G}$), \textit{Treatment} ($\mathbb{T}$), \textit{Diagnosis} ($\mathbb{D}$), \textit{Demographics} ($\mathbb{E}$), and \textit{Disease} ($\mathbb{S}$), defined as follows:

\begin{itemize}[leftmargin=*,noitemsep,nolistsep]
    \item $\mathbb{A} = \{a_1, \ldots, a_q\}$, where $a_i$ takes the form of a numerical constant, or a range. For later comparison purposes, all the integer values within this range are included.
    \item $\mathbb{G} = \{g_1\}$, a singleton set where $g_1$ takes the form of a string constant (e.g., \textit{Female}, \textit{Any}).
    \item $\mathbb{T} = \{t_1, \ldots, t_n\}$, where $t_i$ can take the form of string constants or noun-phrases (e.g., \textit{no prior treatment}) describing patient's treatment history.
    \item $\mathbb{D} = \{d_1, \ldots, d_m\}$, where $d_i$ can take the form of a string constant or noun phrase (e.g., \textit{Age-related macular degeneration}) describing the patient diagnosis.
    \item $\mathbb{E} = \{e_1, \ldots, e_p\}$, where $e_i$ can take the form of a string or noun phrase (e.g., \textit{family history of eye condition}) describing a demographic attribute, other than age and gender.
    \item $\mathbb{S} = \{s_1, \ldots, s_h\}$, where $s_i$ can take the form of a string or noun phrase that denotes a sign or symptom (e.g., \textit{worsening vision in both eyes}).
\end{itemize}

\subsubsection{Attribute extraction}

Attributes \textit{Diagnosis, Demographics, Treatment} and \textit{Disease} are extracted simultaneously via LLM prompting. Firstly, natural language noun phrases are extracted from the patient note. These extracted propositions are then associated with one of the following categories: \textit{Demographics, Treatment}, and \textit{Disease}. Then the LLM is re-prompted to postulate \textit{Diagnosis} values by considering the prevalence of potential conditions and the distinctiveness of the described symptoms. Each attribute type preserves the context and meaning while restructuring the patient note into simpler, self-contained representation, facilitating isolated inference and assessment of each characteristic’s significance.

Attributes \textit{Age} and \textit{Gender} are treated as subsets of the much wider \textit{Demographics} set. In practice, their values are extracted by locating age- and gender-specific terms within the elements of the \textit{Demographics} set and standardizing the output to a single value, a numerical inequality or range (in the case of age).

\begin{figure}[t]
\centering
\includegraphics[width=0.9\columnwidth]{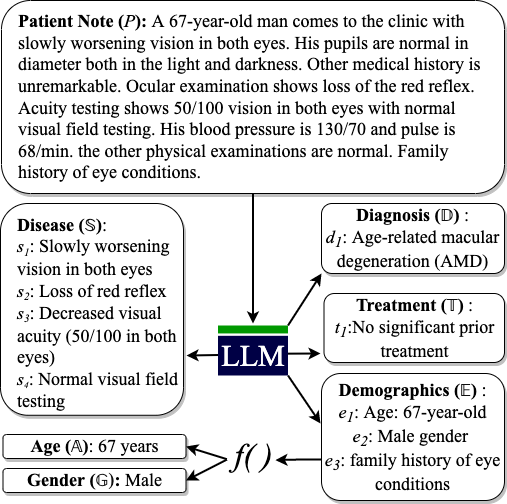}

\caption{Diagram of the patient note structure}
\label{fig:patient_struct}
\end{figure}

Figure \ref{fig:patient_struct} illustrates the overall process described above. Give a patient note, a LLM is prompted for the extraction of text spans, numbers or expressions that constitute the value-set of the six attributes introduced above.

\subsection{CTR modeling} \label{subsec: CTR model}

Following a similar approach, we formally define a clinical trial record $R$ as a collection of attributes \textit{Age} ($\mathbb{A}$), \textit{Gender} ($\mathbb{G}$), \textit{Treatment} ($\mathbb{T}$), \textit{Demographics} ($\mathbb{E}$), \textit{Disease} ($\mathbb{S}$), and \textit{Condition} ($\mathbb{C}$). The value-set of the first five attributes (\textit{viz.} $\mathbb{A}$, $\mathbb{G}$, $\mathbb{T}$, $\mathbb{E}$, and $\mathbb{S}$) take the form of their patient note counterparts (the distinction between them being clear from context). \textit{Condition}, corresponds to a patient diagnosis $\mathbb{D}$ and its value set is defined as $\mathbb{C} = \{c_1, \ldots c_q\}$, where $c_i$ denotes a trial's targeted condition.

\subsubsection{Attribute extraction} \label{subsec: CTR extract}

Attributes \textit{Demographics}, \textit{Treatment}, and \textit{Disease} are extracted via LLM-prompting for individual eligibility criterion categorization\footnote{CTR sources such as \url{https://www.clinicaltrials.gov/} often structure CTR data such that the identification of inclusion/exclusion criteria is trivial.}. 

The extraction of \textit{Condition}, \textit{Age} and \textit{Gender} relies on the CTR's original structure that clearly mentions the trial's targeted condition and by locating age- and gender-specific tokens in the CTR's content, similar to patient note extraction.

\begin{figure}[t]
\centering
\includegraphics[width=\columnwidth]{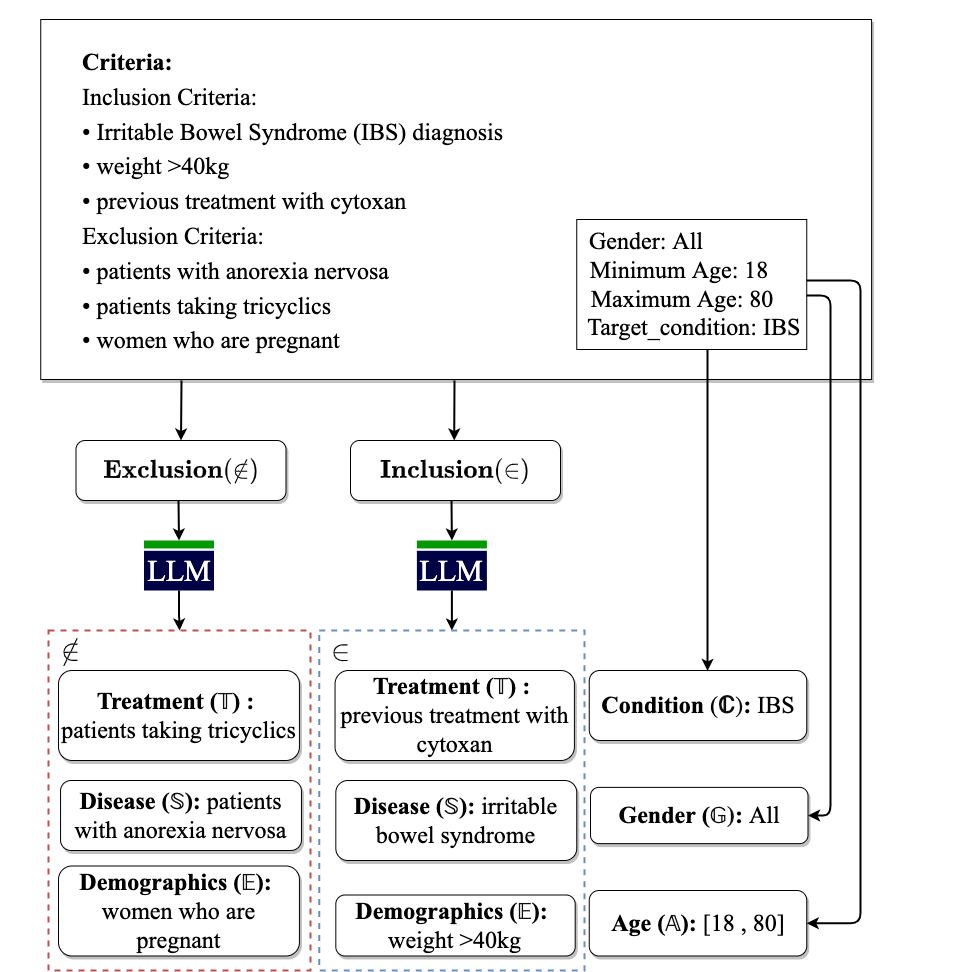}

\caption{Diagram of the CTR structure}
\label{fig:CTR_model}
\end{figure}

Figure \ref{fig:CTR_model} illustrates the overall CTR modeling process described above. Given a CTR, a LLM is prompted for the extraction of text spans and numerical expressions constituting the value-sets of the attributes introduced in this section.

\subsection{Attribute similarity-based normalization \& expansion}
\label{at-norm}
Having modeled the defining characteristics of patients and CTRs as sets, it becomes apparent that one approach to reason about the eligibility of a given patient $P$ to a given trial $R$ is by means of set algebra. However, matching categorical or noun-phrase values often requires fuzzy- and semantic-matching techniques \cite{milajevs2015ir} due to the variations inherent to the medical nomenclature. In this section, we formalize a similarity-based approach to first normalize the attribute values with respect to a domain reference, and then to expand them, when relevant, to a set of candidate values, such that to enable and maximize downstream matching likelihood. In this process, we focus on patient \textit{Diagnosis} and trial targeted \textit{Condition}, as these attributes bear the significant alignment signal.

\subsubsection{Attribute normalization}

Given $\mathbb{D}^P = \{d_1, \ldots, d_m\}$, the set of diagnosis values for a patient $P$, and $\mathbb{C}^R = \{c_1, \ldots c_q\}$, the set of condition values of a CTR $R$, for each $d_i \in \mathbb{D}^P$ and $c_j \in \mathbb{C}^R$ we define their normalized variants $\widehat{d_i} = argmax_{t \in \mathcal{O}}~sim(d_i, t)$ and $\widehat{c_j} = argmax_{t \in \mathcal{O}}~sim(c_j, t)$, where $\mathcal{O}$ is a reference ontology for the medical domain (e.g., Systematized Nomenclature of Medicine-Clinical Terms (SNOMED CT) \cite{donnelly2006snomed}). $\mathcal{O}$ consists of standardized and universal representation of concepts (denoted by $t$), properties, and relationships between concepts within the domain, organized in a taxonomic structure. $sim$ denotes a similarity function (e.g., Jaccard) between an attribute value and some ontology concept $t$. In practice, given the potential size of the concept-set in the ontology, a nearest-neighbour search algorithm (e.g., Locality Sensitive Hashing (LSH) \cite{datar2004locality}) could be employed to efficiently normalize each noun-phrase or constant $d_i/c_j$ to their most similar terms $t_i/t_j \in \mathcal{O}$. Finally, the normalized $\mathbb{D}^P$ and $\mathbb{C}^R$ are defined by: $\widehat{\mathbb{D}^P} = \{\widehat{d_1}, \ldots, \widehat{d_m}\}$ and $\widehat{\mathbb{C}^R} = \{\widehat{c_1}, \ldots \widehat{c_q}\}$, respectively.

\subsubsection{Diagnosis expansion}

\begin{figure}[t]
\centering
\includegraphics[width=\columnwidth]{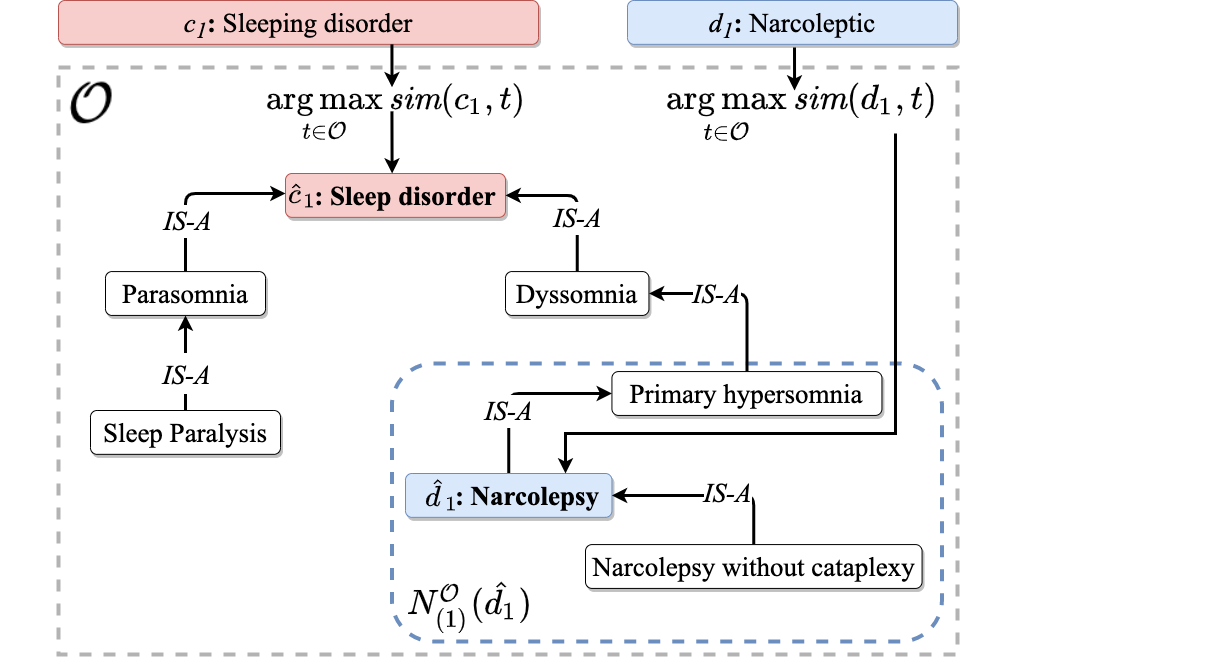}

\caption{Diagram of 1-level relevance Ontology Retrieval. In this example, the target condition falls outside the 1-level relevance of the diagnosis.}
\label{fig:onto}
\end{figure}

In addition to normalization, the domain ontology opens the possibility for leveraging its defined properties and hierarchical concept relationships to expand some of the normalized attribute values with their relevant ontological neighborhood. To define the ontological neighborhood $N^{\mathcal{O}}(t)$ of some concept $t \in \mathcal{O}$, consider the sequence of $is-a$ relationships $(\iota_1 \ldots \iota_n)$ in the ontology between two connected concepts $t_1, t_{n+1}$ \cite{zheng2020missing}. If there is a sequence of terms $(t_1 \ldots t_{n+1})$, such that $\iota_i \doteq t_i\ \text{\textit{is-a}}\ t_{i+1}$ and $t \in (t_1 \ldots t_{n+1})$, then the sequence of concepts defines $N^{\mathcal{O}}_{(n)}(t)$, denoted the $n$-level relevance of $t$. Figure \ref{fig:onto} illustrates the $1$-level relevance of term \textit{Narcolepsy}.

We apply the predicate expansion method to a patient normalized diagnosis, $\widehat{\mathbb{D}^P}$, to maximize its match against a trial's targeted conditions. Thus, the expansion of normalized $\widehat{\mathbb{D}^P}$ is defined by $\widetilde{\mathbb{D}^P} = \bigcup\limits_{i=1}^{m} N^{\mathcal{O}}_{(n)}(\widehat{d_i})$.

\section{A set reasoning framework for clinical trial retrieval and re-ranking}

Following the notations introduced in the previous section, a patient note $P$ can now be defined by six attribute sets, as follows: $\mathbb{A}^P = \{a^P_1, \ldots, a^P_q\}$, $\mathbb{G}^P = \{g^P_1\}$, $\mathbb{T}^P = \{t^P_1, \ldots, t^P_n\}$, $\widetilde{\mathbb{D}^P} = \bigcup\limits_{i=1}^{m} N^{\mathcal{O}}_{(n)}(\widehat{d^P_i})$, $\mathbb{E}^P = \{e^P_1, \ldots, e^P_p\}$, $\mathbb{S}^P = \{s^P_1, \ldots, s^P_h\}$.

Similarly, a CTR can now be defined by six attribute sets, as follows: $\mathbb{A}^R = \{a^R_1, \ldots, a^R_q\}$, $\mathbb{G}^R = \{g^R_1\}$, $\mathbb{T}^R = \{t^R_1, \ldots, t^R_n\}$, $\widehat{\mathbb{C}^R} = \{\widehat{c_1}, \ldots \widehat{c_q}\}$, $\mathbb{E}^R = \{e^R_1, \ldots, e^R_p\}$, $\mathbb{S}^R = \{s^R_1, \ldots, s^R_h\}$. Recall from section \ref{subsec: CTR extract} that the values of $\mathbb{T}^R$, $\mathbb{E}^R$, and $\mathbb{S}^R$ are derived via LLM-prompting from $R$'s inclusion and exclusion criteria. In order to preserve the values' inclusion/exclusion semantics, we further split the mentioned sets into $({^{\in}}\mathbb{T}^R, {^{\notin}}\mathbb{T}^R)$, $({^{\in}}\mathbb{E}^R, {^{\notin}}\mathbb{E}^R)$, and $({^{\in}}\mathbb{S}^R, {^{\notin}}\mathbb{S}^R)$, where ${^{\in}}\mathbb{T}^R \cup {^{\notin}}\mathbb{T}^R = \mathbb{T}^R$, ${^{\in}}\mathbb{E}^R \cup {^{\notin}}\mathbb{E}^R = \mathbb{E}^R$, and ${^{\in}}\mathbb{S}^R \cup {^{\notin}}\mathbb{S}^R = \mathbb{S}^R$, respectively. Intuitively, each element of a $\in$ set denotes an attribute value derived from some inclusion criterion, while each element of a $\notin$ set denotes an attribute value derived from some exclusion criteria.

\subsection{Clinical trial retrieval: from set-definitions to relevance}
\label{c-relevance}
Commonly, state-of-the-art \cite{Kusa2023EffectiveMO} addresses the clinical trial retrieval problem as a two-stage process consisting of an initial ranked candidate retrieval, and a subsequent re-ranking task that promotes the best matches to the top of the ranking. In this paper we build-upon this paradigm of retrieval and re-ranking. 

Building on the attribute-set formalization introduced in the previous section, we define the notion of \textit{relevance} of a CTR $R$ for a given patient $P$:

\begin{definition}[Relevance]
\label{def:relevance}
    The relevance of some CTR $R$ with respect to some patient note $P$, denoted by $\Gamma^P(R)$, is an attribute of the CTR indicating that $R$ targets at least one condition that matches $P$'s diagnosis ($\Gamma_{\mathbb{C}}^P(R)$), and the age/gender conditions of $R$ match the patient's age/gender ($\Gamma_{\mathbb{A}}^P(R)$/$\Gamma_{\mathbb{G}}^P(R)$). 
\end{definition}

We formally define relevance with respect to $\mathbb{C}$, $\Gamma_{\mathbb{C}}^P(R)$, given a patient note $P$ with its diagnostic value-set $\widetilde{\mathbb{D}^P} = \bigcup\limits_{i=1}^{m} N^{\mathcal{O}}_{(n)}(\widehat{d^P_i})$, and a CTR $R$ with its normalized targeted treatment value-set $\widehat{\mathbb{C}^R} = \{\widehat{c_1}, \ldots \widehat{c_q}\}$, as 
\begin{align}
\label{eq:condition-relevance}
\Gamma_{\mathbb{C}}^P(R) = \widetilde{\mathbb{D}^P} \cap \widehat{\mathbb{C}^R}    
\end{align}

\subsection{Clinical trial filtering}

Additionally, the approach proposed in this work allows for additional, albeit weaker, relevance types, \textit{viz.}, age-relevance ($\Gamma_{\mathbb{A}}^P(R)$) and gender-relevance ($\Gamma_{\mathbb{G}}^P(R)$). Specifically, given a patient note $P$ with associated age and gender attribute-sets, $\mathbb{A}^P = \{a^P_1, \ldots, a^P_q\}$ and $\mathbb{G}^P = \{g^P_1\}$, and a CTR $R$ with its age and gender attribute-sets, $\mathbb{A}^R = \{a^R_1, \ldots, a^R_q\}$, $\mathbb{G}^R = \{g^R_1\}$

\begin{align}
\label{eq:age-gen-relevance}
    & \Gamma_{\mathbb{A}}^P(R) = \mathbb{A}^P \cap \mathbb{A}^R \\ 
    & \Gamma_{\mathbb{G}}^P(R) = \mathbb{G}^P \cap (\mathbb{G}^R \cup ``\texttt{All}")
\end{align}

where $``\texttt{All}"$ denotes any case variation thereof. In practice, $\Gamma_{\mathbb{A}}^P(R)$ and $\Gamma_{\mathbb{G}}^P(R)$ can be used as a demographic filter (DF), applied on condition-relevant candidates.

\subsection{Initial clinical trial retrieval}
\label{subsec:initial_ret}

Having defined the age-/gender-/condition-relevance notions of a CTR, given a patient note, we treat the concept of condition relevance as a means for initial retrieval of clinical trials, followed by age and gender filtering, as defined above. In other words, the higher the overlap between a CTR's condition and a patient's diagnosis, the more relevant the trial would be. We refer to this overlap as $ov_{\mathbb{C}}(R)$ given by:

\begin{align}
\label{eq:overlap}
    ov_{\mathbb{C}}(R) = \frac{\left| \Gamma^P_{\mathbb{C}}(R) \right|}{min(\left| \widetilde{\mathbb{D}^P} \right|, \left| \widehat{\mathbb{C}^R} \right|)}
\end{align}
or the \textit{overlap coefficient} between expanded condition values and normalized diagnosis set. In practice, given the potential size of the trial retrieval space (e.g. 375,581 CTRs in the TREC 2022 snapshot \cite{Roberts2022OverviewOT}), information retrieval algorithms, such as LSH or BM25 \cite{robertson1976relevance}, can efficiently quantify the $\mathbb{C}$-relevance of a CTR, since such similarity approximation algorithms return scores that are positively correlated with the overlap coefficient.

\subsection{Clinical trial re-ranking: from relevance to eligibility}

Given a patient note $P$ and its associated attribute-sets, applying the reasoning process described above against a collection of CTRs results in a relevance-ranked result-set $\mathcal{R} = \{R_1, \ldots R_K\}$, i.e., the most condition-/age-/gender-relevant $K$ CTRs, with the ranking given by the above-defined $ov_{\mathbb{C}}(R)$ of CTRs that have also passed the age/gender filtering. Next, we perform further \textit{eligibility} (as opposed to just relevance) analysis by means of labeling. 

\begin{definition}[Eligibility]
\label{def:eligbility}
    The eligibility of some CTR $R$ with respect to some patient note $P$, denoted by $\Xi^{P}(R)$, is an attribute of the CTR indicating that its inclusion/exclusion criteria allow the patient's participation in the study.
\end{definition}

We formally define eligibility with respect to inclusion/exclusion treatment (${^{\in}}\Xi^P_{\mathbb{T}}$/${^{\notin}}\Xi^P_{\mathbb{T}}$), inclusion/exclusion demographics (${^{\in}}\Xi^P_{\mathbb{E}}$/${^{\notin}}\Xi^P_{\mathbb{E}}$), and inclusion/exclusion disease (${^{\in}}\Xi^P_{\mathbb{S}}$/${^{\notin}}\Xi^P_{\mathbb{S}}$). In each case, we employ a labeling function $\phi(v, \mathfrak{c}_{\{\cdot\}^P}) = L \in \{\mathsf{eligible}, \mathsf{excluded}, \mathsf{not\ enough\ info} \}$, implemented via a LLM-based instruction, that takes as input a natural language description of some attribute value $v$ of $R$ and additional instructive information $\mathfrak{c}_{\{\cdot\}^P}$ (i.e., a natural language description of the patient's attribute values, input and output instructions, and optional representative examples) to label each $R$ with one or more labels based on its $\mathbb{T}, \mathbb{E}, \mathbb{S}$ attributes, as described next.

\subsubsection{Fine-grained labeling}

Considering a candidate trial record, $R \in \mathcal{R}$, the $\phi$-based eligibility reasoning is governed by the following attribute-specific tuples:

\begin{align}
\label{eq:TES-eligibility}
\begin{split}
    & {^{\in}}\Xi_{\mathbb{T}}^{P}(R) = (\phi(t_i^R, \mathfrak{c}_{\mathbb{T}^P})\ |\ t_i^R \in {^{\in}}\mathbb{T}^R) \\
    & {^{\notin}}\Xi_{\mathbb{T}}^{P}(R) = (\phi(t_i^R, \mathfrak{c}_{\mathbb{T}^P})\ |\ t_i^R \in {^{\notin}}\mathbb{T}^R) \\
    & {^{\in}}\Xi_{\mathbb{E}}^{P}(R) = (\phi(e_j^R, \mathfrak{c}_{\mathbb{E}^P})\ |\ e_j^R \in {^{\in}}\mathbb{E}^R)  \\
    & {^{\notin}}\Xi_{\mathbb{E}}^{P}(R) = (\phi(e_j^R, \mathfrak{c}_{\mathbb{E}^P})\ |\ e_j^R \in {^{\notin}}\mathbb{E}^R)  \\
    & {^{\in}}\Xi_{\mathbb{S}}^{P}(R) = (\phi(s_l^R, \mathfrak{c}_{\mathbb{S}^P})\ |\ s_l^R \in {^{\in}}\mathbb{S}^R) \\
    & {^{\notin}}\Xi_{\mathbb{S}}^{P}(R) = (\phi(s_l^R, \mathfrak{c}_{\mathbb{S}^P})\ |\ s_l^R \in {^{\notin}}\mathbb{S}^R)
\end{split}
\end{align} 
defined over an underlying set $\{\mathsf{eligible}, \mathsf{excluded}, \mathsf{not\ enough\ info} \}$.

\subsubsection{Coarse-grained labeling}

By extension, we use use the same labeling function in the form $\phi((\mathbb{T}^R, \mathbb{E}^R,\mathbb{S}^R), (\mathfrak{c}_{\mathbb{T}^P}, \mathfrak{c}_{\mathbb{E}^P}, \mathfrak{c}_{\mathbb{S}^P})) \in \{\mathsf{eligible}, \mathsf{excluded}\}$ to determine an overall eligibility, given by:
\begin{align}
\label{eq:overTES-eligibility}
\begin{split}
    & \Xi_{(\mathbb{T, E, S})}^{P}(R) = \phi((\mathbb{T}^R, \mathbb{E}^R,\mathbb{S}^R), (\mathfrak{c}_{\mathbb{T}^P}, \mathfrak{c}_{\mathbb{E}^P}, \mathfrak{c}_{\mathbb{S}^P})) \\
\end{split}
\end{align} 

\subsection{Deontic-re-ranking for clinical trials}

The well-defined structuring of patient notes and clinical trials leads to more grounded reasoning for retrieval and LLM prompting, i.e., a controlled LLM reasoning framework that maximizes recall and precision. However, the remaining patient-CTR matching is susceptible to information availability challenges: (i) patient notes can be insufficient for direct logical inference in relation to a CTR's eligibility criteria - that is the information in the patient notes neither logically nor directly contradicts nor confirms every eligibility criterion; (ii) the CTR criteria can insufficiently cover the details contained in patient notes - that is the criteria do not logically and directly contradict or confirm every characteristic in the patient notes; and (iii) there is insufficient information in the patient notes and CTR to support the inference of overall eligibility - i.e. there is a requisite threshold of direct logical confirmation between the patient notes and a CTR for overall eligibility to be possible. 

In light of the above challenges, we argue that part of the clinical trials eligibility matching can be conceptualized with the support of deontic logic, following the concepts outlined in \citet{sep-logic-deontic}. Specifically, based on the attribute-set definitions introduced in this paper and their relevance/eligibility qualities defined in relations \ref{eq:condition-relevance}, \ref{eq:age-gen-relevance}, \ref{eq:TES-eligibility}, \ref{eq:overTES-eligibility}, we explore a wide space of re-ranking scoring functions grounded in the following principles:

\begin{enumerate}[leftmargin=*]
    \item We construe $\mathbb{A}^R$, $\mathbb{G}^R$ and $\mathbb{C}^R$ as exclusive properties for relevance. This leads to the following mandatory conjunctive relations: \\
    \begin{align}
    \label{eq:AGR-exclusive}
    \begin{split}
        & (\Gamma_{\mathbb{A}}^P(R) \neq \varnothing) \wedge (\Gamma_{\mathbb{G}}^P(R) \neq \varnothing) \wedge \\
        & (\Gamma_{\mathbb{C}}^{P}(R) \neq \varnothing)
    \end{split}
    \end{align} 
    \item Exclusion is generally impermissible\footnote{In practice, scoring functions that strictly follow this principle lead to poor rankings. Thus, we apply a more lenient version where exclusion is allowed but penalized - to account for LLM reasoning inaccuracies.} and permissible but sub-optimal if eligibility is indeterminate (i.e., $\mathsf{not\ enough\ info}$) with respect to both inclusion and exclusion criteria. This leads to the following mandatory relation:
    \begin{align}
    \label{eq:exclusion}
    \begin{split}
        & \bigwedge\limits_{\mathbb{X} \in \{\mathbb{T, E, S}\}} \mathsf{excluded} \notin {^{\in}}\Xi_{\mathbb{X}}^{P}(R)\ \wedge \\
        & \bigwedge\limits_{\mathbb{X} \in \{\mathbb{T, E, S}\}} \mathsf{excluded} \notin {^{\notin}}\Xi_{\mathbb{X}}^{P}(R)\ \wedge \\
        & (\mathsf{excluded})  \neq {^{\in}}\Xi_{(\mathbb{{^{\in}}T, {^{\in}}E, {^{\in}}S})}^{P}(R)\ \wedge  \\
        & (\mathsf{excluded})  \neq {^{\notin}}\Xi_{(\mathbb{{^{\notin}}T, {^{\notin}}E, {^{\notin}}S})}^{P}(R)
    \end{split}
    \end{align} 
    \item Inclusion criteria are minimally obligatory and omissible but sub-optimal if eligibility is indeterminate. This leads to the following mandatory relation:
    \begin{align}
    \label{eq:inclusion}
    \begin{split}
        & \bigvee\limits_{\mathbb{X} \in \{\mathbb{T, E, S}\}} \mathsf{eligible} \in {^{\in}}\Xi_{\mathbb{X}}^{P}(R)\ \vee \\
        & \bigvee\limits_{\mathbb{X} \in \{\mathbb{T, E, S}\}} \mathsf{eligible} \in {^{\notin}}\Xi_{\mathbb{X}}^{P}(R)\ \vee \\
        & (\mathsf{eligible}) = {^{\in}}\Xi_{(\mathbb{{^{\in}}T, {^{\in}}E, {^{\in}}S})}^{P}(R)\ \vee \\
        & (\mathsf{excluded}) = {^{\notin}}\Xi_{(\mathbb{{^{\notin}}T, {^{\notin}}E, {^{\notin}}S})}^{P}(R)
    \end{split}
    \end{align}
\end{enumerate}

In other words, given a patient $P$ and a CTR $R \in \mathcal{R}$, $R$ is considered for re-ranking iff $R$ is treatment-/age-/gender-relevant (relation \ref{eq:AGR-exclusive}), the generative function $\phi$ does not identify any exclusion instance in any inclusion/exclusion criteria attributes and it identifies at least one eligibility instance in inclusion/exclusion criteria attributes.

Sub-optimality is implicitly factored into the rankings. Specifically, we accept eligibility values labeled as $\mathsf{not\ enough\ info}$ as long as they comply with principles 1 and 2 mentioned above. However, these labels do not contribute to a higher ranking. Instead, they are only considered when calculating the total number of criteria for computing proportions.

\section{Re-ranking Scoring Functions}
Drawing on the above principles, we devise a spectrum of re-ranking scoring functions that explore various reasoning mechanism for producing a ranked list of CTRs, given a patient note. Every scoring function defined below adheres to principle 1: any CTR that is not condition-, age-, and gender-relevant is discarded. The remaining two principles are captured in the scoring functions at various degrees, with the aid of a count function $\mathtt{count}_{\Xi_{\mathbb{X}}}(l)$, defined over $\mathbb{T}, \mathbb{E}$ or $\mathbb{S}$, that counts the occurrences of some eligibility label $l$.


\begin{table*}[h!]
\tiny
\centering
\begin{tabular}{p{2cm}p{8cm}p{4.5cm}}
\hline
\textbf{Score} & \textbf{Function} & \textbf{Description} \\ \hline

\raggedright{\textbf{Inclusion eligibility}}& 

\[
IE\_score(R) = \frac{\sum\limits_{\mathbb{X} \in \{\mathbb{T, E, S}\}} \mathtt{count}_{{^{\in}}\Xi_{\mathbb{X}}}(\mathsf{eligible})} {\sum\limits_{\mathbb{X} \in \{\mathbb{T, E, S}\}} \left| {^{\in}}\mathbb{X} \right|}
\] & Favors CTRs that are deemed eligible with respect to their fine grained inclusion criteria. Additionally Disease-only, Demographic-only, and Treatment-only eligibility, where CTRs are ranked by Inclusion Eligibility, limited to $\mathbb{X} \in \{\mathbb{T}\}, \{\mathbb{E}\},\text{and }
\{\mathbb{S}\}$ respectively.
\\ \hline 

\raggedright{\textbf{Filtered inclusion eligibility}}& 
\[
FIE\_score(R) = 
\begin{cases} 
0 & \text{if } \sum\limits_{\mathbb{X} \in \{\mathbb{T, E, S}\}} \mathtt{count}_{\Xi_{\mathbb{X}}}(\mathsf{excluded}) > 0 \\
IE\_score(R) & \text{otherwise}
\end{cases}
\] & Inclusion eligibility, discarding any trial candidate that has been labeled excluded in any fine grained criteria. Additionally, a variant of this scoring method is employed, Filtered inclusion-Only Eligibility, which discards any trial candidate that has been labeled as excluded based on the fine-grained inclusion criteria only.

\\ \hline

\raggedright{\textbf{Exclusion eligibility}}& 

\[
EE\_score(R) = \frac{\sum\limits_{\mathbb{X} \in \{\mathbb{T, E, S}\}} \mathtt{count}_{{^{\notin}}\Xi_{\mathbb{X}}}(\mathsf{eligible})} {\sum\limits_{\mathbb{X} \in \{\mathbb{T, E, S}\}} \left| {^{\notin}}\mathbb{X} \right|}
\] &The complementary score of inclusion eligibility with a focus on the granular exclusion criteria.
\\ \hline

\raggedright{\textbf{General eligibility}}& 

\[
GE\_score(R) = \frac{\sum\limits_{\mathbb{X} \in \{\mathbb{T, E, S}\}} \mathtt{count}_{\Xi_{\mathbb{X}}}(\mathsf{eligible})} {\sum\limits_{\mathbb{X} \in \{\mathbb{T, E, S}\}} \left| \mathbb{X} \right|}
\] &Indiscriminately favors eligible CTRs with respect to both fine-grained inclusion and exclusion criteria.
\\ \hline

\raggedright{\textbf{Contrasting eligibility}}& 

\[
\frac{\sum\limits_{\mathbb{X} \in \{\mathbb{T, E, S}\}} \mathtt{count}_{\Xi_{\mathbb{X}}}(\mathsf{eligible}) - \sum\limits_{\mathbb{X} \in \{\mathbb{T, E, S}\}} \mathtt{count}_{\Xi_{\mathbb{X}}}(\mathsf{excluded})}{\sum\limits_{\mathbb{X} \in \{\mathbb{T, E, S}\}} \left| \mathbb{X} \right|}
\] &Quantifies eligibility dominance over non-eligibility favoring CTRs with many eligible properties and few excluding properties.
\\ \hline

\raggedright{\textbf{Weighted contrasting eligibility}}& 

\[
\frac{\alpha \times \sum\limits_{\mathbb{X} \in \{\mathbb{T, E, S}\}} \mathtt{count}_{\Xi_{\mathbb{X}}}(\mathsf{eligible}) - \beta \times \sum\limits_{\mathbb{X} \in \{\mathbb{T, E, S}\}} \mathtt{count}_{\Xi_{\mathbb{X}}}(\mathsf{excluded})}{\sum\limits_{\mathbb{X} \in \{\mathbb{T, E, S}\}} \left| \mathbb{X} \right|}
\]&Considers configurable weights for eligibility and non-eligibility factors of Contrasting eligibility.
\\ \hline

\raggedright{\textbf{Coarse-grained eligibility}}& 

\[
CG\_score(R) = 
\begin{cases} 
1 + ov(R) & \text{if } (\textsf{eligible}) = \Xi_{(\mathbb{T, E, S})}^{P}(R) \\
ov(R) & \text{otherwise}
\end{cases}
\]&Increases the initial ranking overlap by one for eligible attribute values to further favor eligibility.
\\ \hline

\raggedright{\textbf{Hybrid eligibility}}& 
\[
H_score(R) = 
\begin{cases} 
1 + IE\_score(R) & \text{if } (\textsf{eligible}) = \Xi_{(\mathbb{T, E, S})}^{P}(R) \\
IE\_score(R) & \text{otherwise}
\end{cases}
\]&Combines fine and coarse grained evidence by increasing the eligibility score corresponding to the former when the latter exists.
\\ \hline

\end{tabular}
\caption{Formal definitions of scoring functions}
\end{table*}

\begin{table}[h!]
\centering
\scriptsize
\begin{tabular}{p{3.6cm} p{0.8cm} p{0.35cm} p{0.35cm} p{0.35cm}}
\hline
\textbf{Ranking Method} & \textbf{NDCG@10} & \textbf{P@10} & \textbf{P@25} & \textbf{MRR}  \\ \hline
BM25 & 0.275 & 0.342 & 0.302 &  0.555  \\ \hline
BM25-DF & 0.293 & 0.348 & 0.314 & 0.551  \\ \hline
\multicolumn{5}{l}{\textbf{GPT-3.5}} \\ \hline
$ov_{\mathbb{C}}$  & 0.544 & 0.654 & 0.630 & 0.807 \\ \hline
$ov_{\mathbb{C}}$-DF & 0.567 & 0.650 & 0.610 & 0.796 \\ \hline
Filtered inclusion eligibility & 0.443 & 0.466 & 0.239 & 0.706 \\ \hline
Inclusion eligibility & 0.574 & 0.664 & 0.618 & 0.777 \\ \hline
General eligibility  & 0.566 & 0.644 & 0.609 & 0.776 \\ \hline
Contrasting eligibility  & 0.570 & 0.650 & 0.611 & 0.801 \\ \hline
Weighted contrasting eligibility  & 0.561 & 0.628 & 0.594 & 0.789 \\ \hline
Coarse-grained eligibility  & 0.567 & 0.656 & 0.630 & 0.753 \\ \hline
Hybrid eligibility  & 0.632 & 0.702 & 0.648 & 0.827 \\ \hline
\multicolumn{5}{l}{\textbf{GPT-4-turbo}} \\ \hline
$ov_{\mathbb{C}}$ & 0.550 & 0.664 & 0.629 & 0.807 \\ \hline
$ov_{\mathbb{C}}$-DF & 0.575 & 0.654 & 0.606 & 0.807 \\ \hline
Filtered inclusion eligibility & 0.530 & 0.532 & 0.248 & 0.816 \\ \hline
Filtered inclusion-only eligibility & 0.632 & 0.686 & 0.422 & 0.845 \\ \hline
Disease-only eligibility & 0.641 & 0.702 & 0.576 & 0.860 \\ \hline
Demographic-only eligibility & 0.539 & 0.614 & 0.406 & 0.724 \\ \hline
Treatment-only eligibility & 0.447 & 0.534 & 0.273 & 0.720 \\ \hline
Inclusion eligibility & 0.638 & 0.702 & 0.630 & 0.837 \\ \hline
Exclusion eligibility & 0.569 & 0.658 & 0.630 & 0.771 \\ \hline
General eligibility & 0.624 & 0.686 & 0.614 & 0.859 \\ \hline
Contrasting eligibility  & 0.612 & 0.678 & 0.614 & 0.839 \\ \hline
Weighted contrasting eligibility  & 0.628 & 0.680 & 0.630 & 0.845 \\ \hline
Coarse-grained eligibility  & \textbf{0.693} & 0.718 & \textbf{0.631} & 0.769 \\ \hline
Filtered Coarse-grained eligibility & 0.658 & 0.650 & 0.376 & 0.763 \\ \hline
Hybrid eligibility  & 0.679 & \textbf{0.730} & 0.630 & \textbf{0.860} \\ \hline
\textbf{TREC SOTA}  & 0.613 & 0.508 & - & 0.726 \\ \hline
\textbf{TREC Median} & 0.498 & 0.380 & - & 0.598 \\ \hline
\end{tabular}
\caption{Performance metrics of various retrieval methods on the TREC 2022 Clinical trial track.}
\label{results}
\end{table}

We present the findings of our experiments conducted on the TREC 2022 Clinical Trials Track dataset. All evaluations adhered to the official TREC protocols, and we used TrecTools for generating and validating the results shown in Table \ref{results}. 

In our experimental setup, we utilized the SNOMED CT domain ontology, Jaccard similarity, LSH and 1 level $\mathbb{C}$ relevance for attribute normalization REF, and the initial retrieval REF. For the initial ranking we employed BM25 for ($ov_{\mathbb{C}}$) REF. Due to resource constraints, we limited re-ranking to the top 25 CTRs from this initial ranking. In cases where DF output fewer than 25 trials, we re-applied the $ov_{\mathbb{C}}$ ranking and DF to the entire set, augmenting the list with top-ranked trials until we achieved a total of 25, adjusting the rankings such that CTRs sourced through $\mathbb{C}$ relevance were always ranked higher.

\subsection{Initial Retrieval and Ranking}

\begin{table}[h!]
\small
    \centering
    \begin{tabular}{>{\raggedright\arraybackslash}p{2cm} >{\centering\arraybackslash}p{1cm} >{\centering\arraybackslash}p{1cm} >{\centering\arraybackslash}p{1cm}}
        \hline
        \textbf{Method} & \textbf{R@10} & \textbf{R@25} & \textbf{R@500} \\
        \hline
        BM25 & 0.03 & 0.06 & 0.761 \\
        $ov_{\mathbb{C}}$ GPT4 & 0.088 & \textbf{0.204} & \textbf{0.859} \\
        $ov_{\mathbb{C}}$-GPT4-DF & \textbf{0.105} & 0.202 & 0.791 \\
        \hline
    \end{tabular}
    \caption{Recall@N for different initial ranking methods}
    \label{tab:recall_scores}
\end{table}

\paragraph{$ov_{\mathbb{C}}$ outperforms TREC SOTA in P@10 and MRR} 
As shown in Table \ref{results} $ov_{\mathbb{C}}$ outperformed the TREC Median across all metrics, regardless of the underlying LLM used for diagnosis extraction. Additionally, GPT-4 $ov_{\mathbb{C}}$-DF underperforms in NDCG@10 compared to the TREC 2022 SOTA \textit{frocchio monot5 e} by a margin of -0.038, but significantly outperforms the SOTA with improvements of +0.146 in P@10 and +0.081 in MRR. 

\paragraph{1 level $\mathbb{C}$ relevance retrieval significantly outperforms the baseline in Recall} As outlined in Table \ref{tab:recall_scores}, GPT-4 $ov_{\mathbb{C}}$ significantly outperformed BM25 in Recall @10, @25, and @500, by 0.058, 0.144, and 0.098 respectively. 

\paragraph{Diminishing Returns in Higher \textit{n}-level $\mathbb{C}$ Relevance}
A study was conducted to assess the performance of $\mathbb{C}$ relevance at various \textit{n}-levels, with the results shown in Table \ref{tab:relevance}. The analysis revealed a point of diminishing returns as \textit{n} increases, particularly between levels 3 and 4. While recall improved slightly by +0.034, precision dropped significantly by -0.102. This aligns with the expectation that increasing ontological taxonomic distance correlates with reduced relevance. However, the findings also indicate that the majority of relevant clinical trials are located within 3 levels of relevance in the ontology, making further exploration less effective for practical retrieval.

\paragraph{Ontological attribute normalisation is robust}
The normalization of diagnoses and conditions (REF) to the SNOMED-CT ontology is robust, achieving an average Jaccard similarity score of 0.812 between original diagnoses and conditions and their normalized variants. 

\paragraph{LLMs demonstrates robust diagnostic capabilities within the TREC setting} Several studies have noted the difficulty LLMs encounter in producing accurate differential diagnoses, often resulting in poor performance \cite{omar2024utilizing}. The assessment of GPT-4 Turbo’s diagnostic capabilities within TREC 2022 involved a survey conducted by a medical professional, evaluating its diagnoses across 50 patient topics in the TREC dataset. The survey categorized diagnoses as correct, plausible, doubtful, or incorrect. See Figure \ref{survey} in the appendix for a sample of this survey. The evaluation results indicate that GPT-4 Turbo produced 48/50 correct diagnoses, with all 50 diagnoses classified as plausible. Notably, among the 50 patient notes, 24 either explicitly provided or suggested a specific diagnosis, all of which were accurately identified by GPT-4 Turbo. 

\paragraph{$\mathbb{C}$ relevance is an effective  model of clinical relevance} As established $\mathbb{C}$ relevance has a robust foundation of accurate diagnosis extraction, and normalization of diagnoses and conditions, that preserves the semantic integrity of the original terms. Allowing the $ov_{\mathbb{C}}$ rankings of the $\mathbb{C}$ relevance retrieval to to outperform SOTA methods in precision and mean reciprocal rank (MRR), while also surpassing baseline models in recall across all levels. This is particularly impressive as these metrics assess the ability to identify eligibility, not just relevance. This demonstrates that the definition of $\mathbb{C}$ relevance (\ref{def:relevance}) is strongly correlated with eligibility.

\begin{table}[h!]
\small
    \centering
    \begin{tabular}{>{\centering\arraybackslash}p{0.5cm} >{\centering\arraybackslash}p{0.6cm} >{\centering\arraybackslash}p{1cm} >{\centering\arraybackslash}p{3.5cm}}
        \hline
        \textbf{Level} & \textbf{Recall} & \textbf{Precision} & \textbf{Mean CTRs per Patient} \\
        \hline
        1 & 0.433 & \textbf{0.741} & 81.54 \\
        2 & 0.491 & 0.601 & 113.94 \\
        3 & 0.625 & 0.412 & 211.58 \\
        4 & \textbf{0.659} & 0.310 & 296.32 \\
        \hline
    \end{tabular}
    \caption{Results for OR with different level of relevance}
    \label{tab:relevance}
\end{table}

\begin{figure}[h!]
\centering
\includegraphics[width=\columnwidth]{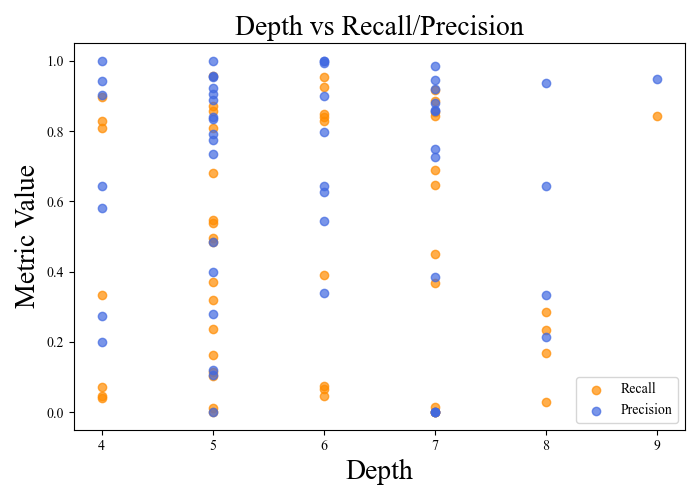}
\caption{Graph of Recall and Precision of 1 level $\mathbb{C}$ Relevance Retrieval by Depth of Diagnosis.}
\label{fig:recall g}
\end{figure}

\paragraph{$\mathbb{C}$ Relevance is Agnostic to Diagnosis Specificity}
A study was conducted to analyze the precision and recall of 1-level $\mathbb{C}$ relevance retrieval based on the depth\footnote{Depth refers to the number of \textit{is-a} relationships from some $t \in \mathcal{O}$ to the root of the SNOMED-CT ontology.} of a normalized diagnosis within the SNOMED-CT ontology, as shown in Figure \ref{fig:recall g}. It was hypothesized that more specific conditions, such as non-small cell lung cancer, would yield higher precision due to being targeted by fewer trials, while broader diagnoses, like lung cancer, would have higher recall. However, results showed significant variation in precision and recall for diagnoses of the same depth. Pearson correlation coefficients for recall and precision with respect to depth were -0.015 and -0.021, respectively, indicating that diagnosis specificity does not significantly impact the performance of $\mathbb{C}$ relevance retrieval.

\subsubsection{Evaluation of Demographic Filtering}
The integration of DF consistently improved the NDCG@10 scores, with increases of +0.018, +0.023, and +0.025 for BM25, $ov_{\mathbb{C}}$ with GPT-3.5, and $ov_{\mathbb{C}}$ with GPT-4 Turbo, respectively, as documented in Table \ref{results}. Conversely, this method has resulted in average reductions of -0.0027, -0.0103, and -0.005 in P@10, P@25, and MRR, respectively. For GPT-4 $ov_{\mathbb{C}}$, DF produced +0.017 in Recall@10, with decreases of -0.068 and -0.002 in Recall@25 and Recall@500, respectively, as outlined in Table \ref{tab:recall_scores}.

Furthermore, the implementation of DF on the $ov_{\mathbb{C}}$ GPT-4 retrieval @500 led to the exclusion of an average of 115.92 CTR per topic from the retrieval pool, achieving a precision of 98\% in eliminating trials that are $\mathsf{not relevant}$ or $\mathsf{excluded}$.

DF incorrectly discards 33 $\mathsf{eligible}$ trials from the top 25 $ov_{\mathbb{C}}$ GPT-4 retrieval, across all 50 patients. A manual examination of these trials showed that 6/33 of these trials were misclassified. For instance, in the case of a \textit{“38-year-old man”} (Patient note 46), Trials NCT04483830 and NCT04652765, both stipulating a minimum participant age of 40, were erroneously annotated as $\mathsf{eligible}$. However, the majority of misclassifications can be attributed to inaccurate extractions of age values and units from GPT-4 Turbo outputs.

\subsection{Re-Ranking}

\paragraph{Comparison with SOTA:}
GPT-4 Inclusion eligibility, General eligibility, Coarse-grained eligibility, and Hybrid eligibility all significantly outperform the current TREC SOTA across all reported metrics. Due to the lack of experimental details, it is not feasible to further compare our approach with \textit{frocchio monot5 e}. GPT-4 Hybrid eligibility recorded the highest P@10 of 0.73 and MRR of 0.86, whereas Coarse-grained eligibility reported the highest NDCG@10 of 0.693 and P@25 of 0.631.

\paragraph{Coarse-grained eligibility outperforms fine-grained re-ranking}
As shown in Table \ref{results}, GPT-4 Coarse-grained Eligibility outperforms the best fine-grained re-ranking with +0.052 in NDCG@10, +0.016 in P@10, and +0.001 in P@25, though it underperforms by -0.091 in MRR. While GPT-3.5 Coarse-grained Eligibility underperformed across all metrics.

\paragraph{Inclusion of Exclusion Criteria Labels Degrades Performance}
As shown in Table \ref{results}, scoring functions underperform when leveraging exclusion criteria labels. Specifically, Exclusion Eligibility, which relies solely on exclusion criteria labels, exhibits a performance drop of -0.006 in NDCG@10 and -0.036 in MRR in comparison to $ov_{\mathbb{C}}$-DF . Moreover, it is outperformed by Inclusion Eligibility by margins of +0.069 in NDCG@10, +0.044 in P@10, and +0.066 in MRR. Similarly, the top-performing re-ranking method incorporating exclusion criteria labels, Weighted Contrastive Eligibility, lags behind Inclusion Eligibility by -0.010 in NDCG@10, -0.022 in P@10, -0.037 in P@25, and -0.074 in MRR. Furthermore, the GPT-4 Filtered Inclusion-only Eligibility demonstrates notable improvements, outperforming Filtered Inclusion Eligibility by 0.102 in NDCG@10, 0.154 in P@10, 0.174 in P@25, and 0.029 in MRR. These patterns are consistent with GPT-3.5. The evidence indicates a clear trend: the inclusion of exclusion criteria labels leads to significant performance degradation, underscoring the substantial discrepancy in LLM performance between labeling exclusion versus inclusion criteria.

\paragraph{Exclusion Criteria Require More Complex Logical Reasoning Than Inclusion Criteria}

We hypothesize several reasons for the observed complexity of exclusion criteria. The primary challenge lies in their inherently negated nature, where satisfying a criterion involves the absence of a condition. Negation, as a logical construct, has been shown to pose challenges even for SOTA LLMs \cite{truong2023language}. Furthermore, exclusion criteria frequently include double negatives (e.g., \textit{“unable to…”} or \textit{“no presence of…”}). This pattern, often mixed with single negation criteria, introduces inconsistency and adds complexity. In contrast, inclusion criteria are typically straightforward assertions.

Therefore solving inclusion criteria usually involves proving the presence of a condition, relying on direct, measurable observations. However, for exclusion criteria, proving absence is required, which involves ruling out all possibilities. This process is exhaustive and often incomplete, as the absence of evidence is not necessarily evidence of absence. While negated inclusion criteria do exist (e.g., the exclusion of a specific disease), they are generally more focused, targeting a specific condition. In contrast, exclusion criteria often encompass broader categories, designed to catch edge cases, which further complicates their application and verification.

\paragraph{LLMs can support strict deontic principles}
As shown in Table \ref{results}, the scoring function based on an absolute interpretation of deontic principles, GPT-4-turbo Filtered Inclusion Eligibility, exhibited a reduction of -0.045 in NDCG@10, -0.122 in P@10, and -0.358 in P@25 compared to $ov_{\mathbb{C}}$-DF. Despite these reductions, it significantly outperformed the TREC Median, with improvements of +0.134 in NDCG@10, +0.306 in P@10, and +0.247 in MRR. When accounting for issues related to exclusion criteria, the Filtered Inclusion-Only Eligibility approach outperforms the TREC SOTA, showing gains of +0.019, +0.178, and +0.119 in NDCG@10, P@10, and MRR, respectively. It is important to note that deontic principles are designed not to maximize outcomes at any cost but rather to define transparent, principled intentions and justifications for decisions—\textit{what ought to be done}. Given these findings, we conclude that LLMs can effectively support the enforcement of strict deontic principles, offering a principled approach to decision-making in retrieval tasks.

\paragraph{Disease criteria are the most disciminative attribute for ranking CTRs} Three experiments are conducted to evaluate the discriminative power of the different criteria types. Shown in Table \ref{results}, these are Disease-only, Demographic-only, and Treatment-only eligibility, where CTRs are ranked by Inclusion Eligibility, limited to criteria specific to these categories. The Disease criteria re-ranking outperforms the TREC SOTA across all metrics. Conversely, the performance of Demographic-only and Treatment-only rankings fell below baseline levels, with Treatment criteria demonstrating the weakest discriminative capability. We hypothesise that this is partially due to the substantial overlap of Demographic criteria with DF, and the limited presence of Treatment criteria.

\paragraph{Coarse and Fine-Grained Labels Are Not Aligned}
A comparative analysis was conducted between the outputs of the GPT-4 Filtered Inclusion Eligibility and Filtered Coarse-Grained Eligibility models. While both scoring methods theoretically adhere to the same deontic principles, the inter-annotator agreement, measured by Cohen’s Kappa, yielded a score of 0.208, indicating very limited alignment. This suggests that despite alignment in task, input, and underlying principles, the LLM is producing significantly different outcomes, likely employing distinct reasoning processes. These findings highlight the challenges in understanding how LLMs execute specific tasks and whether they are truly adhering to the intended objectives.

\section{Related work} \label{sec:rel_work}
The domain of clinical trial retrieval and patient matching has gathered significant attention, particularly through contributions made to the TREC Clinical Trials Tracks 2022 \cite{roberts2022overview} and 2021 \cite{soboroff2021overview}. We present the results of various systems applied to the TREC Clinical Trials Tracks 2022 in Table \ref{tab:trec_methods} where we note that a predominant trend is the use of BM25, a well-established probabilistic retrieval model, as the core retrieval mechanism in all but one of the works. 

Mechanisms for improving retrieval performance are often necessary when using BM25. Works such as \citet{peikos2023utilizing} and \citet{Nunzio2022SummarizeAE} use query expansion techniques such as RM3 to enhance the initial query representation and improve the initial retrieval. The top 2 approaches in Table \ref{tab:trec_methods} also integrate demographic and lifestyle filtering to refine retrieval results based on patient characteristics \cite{Kusa2023EffectiveMO, Herrmannova2022ElsevierDS}.

Similar to our method, a number of approaches \cite{peikos2023utilizing, Wu2022CogStackCA, Sin2022JBNUAT} rely on domain specific knowledge but their focus is on refining query representation. Such methods either fine-tuned BERT-based models, such as  ClinicalBERT, BioBERT, and ChatGPT, or use tools like ScispaCy, MedspaCy \cite{Kusa2023EffectiveMO}, and the Medical Concept Annotation Tool \cite{Wu2022CogStackCA} which links to SNOMED-CT.

The integration of advanced ranking models alongside traditional retrieval methods is another notable trend. Fine-tuned BERT-based models, such as BioBERT \cite{Kusa2023EffectiveMO}, MonoBERT \cite{Nguyen2022MatchingAP}, and MiniLM \cite{Herrmannova2022ElsevierDS}, are used to re-rank initial BM25 results, leveraging the deep contextual understanding of transformer-based architectures.

With more focus on the recent advancements in LLM research, the current state-of-the-art approach for clinical trial retrieval, as documented by \citet{Roberts2022OverviewOT}, is \textit{frocchio monot5 e}. However, detailed information about this approach is not available. Another significant contribution is TrialGPT \cite{jin2023matching}, 
a method that uses GPT-4 to assess the patient’s eligibility at individual criterion level, re-prompting the LLM to evaluate the overall eligibility, and aggregating the outputs to determine a rank. TrialGPT focuses on second-stage retrieval and ranking, with a reduced search space. This is different from the state-of-the-art benchmarks, such as the TREC 2022 task, where end-to-end retrieval and ranking methods are evaluated on a search space of more than 35,000 studies. Thus, the results reported by TrialGPT are not directly comparable.

The approach presented in this paper is an end-to-end process that adheres to the TREC specifications by employing a first-stage retrieval built on BM25, ontology and LLM-based structuring, followed by CT criteria filtering and deontic and LLM-based re-ranking. Crucially, and in contrast to the state-of-the-art in LLM-based CT retrieval, this approach emphasizes the grounding and control of LLM generation, together with the inclusion of domain-relevant explanatory elements in the final ranking, both valuable features in the clinical setting.


\begin{table*}[t]
\scriptsize
\centering
\begin{tabular}{p{1.2cm}p{1.7cm}p{1.5cm}p{1.1cm}p{2.2cm}p{4.2cm}p{1.2cm}}

\textbf{Paper} & \raggedright{\textbf{Retrieval}} & \textbf{Ranking model} & \raggedright{\textbf{Other IR Tools}} & \raggedright{\textbf{Final Query representation}}& \textbf{Approach Summary}&\textbf{NDCG@10} \\
\hline
\raggedright{\cite{Kusa2023EffectiveMO}} & BM25, TF-IDF, DFR. & BioBERT & ScispaCy, MedspaCy & \raggedright{Keyword augmented topic} & BM25 TF-IDF retrieval, rule-based Demographic \& lifestyle Filtering, followed by fine-tuned BioBERT re-ranking &0.557\\
\hline
\raggedright{\cite{peikos2023utilizing}} & PyTerrier BM25 & PyTerrier BM25 & RM3 & Keyword-based Query& Negation Removal and ChatGPT Information Extraction, Medical Role
\& Task Description, with RM3 Query expansion& 0.509\\
\hline
\raggedright{\cite{Nunzio2022SummarizeAE}} & PyTerrier BM25 & PyTerrier BM25 & \raggedright{RM3, KS} &  Keyword-based query & Keyword topic summary, RM3 expansion, demographic filtering, \& BM25 re-ranking &0.505\\
\hline
\raggedright{\cite{Nguyen2022MatchingAP}} & BM25 & MonoBERT, SciBERT & None & Unprocessed topic & BM25 retrieval, fine-tuned MonoBERT-style reranker with SciBERT checkpoint &0.491\\
\hline
\raggedright{\cite{Herrmannova2022ElsevierDS}} & BM25 & \raggedright{MiniLM-L6-v2} & None & \raggedright{MiniLM-L6-v2 embeddings}& Age/gender filtering, BM25 retrieval and MiniLM-L6-v2 embedding ranking &0.476\\
\hline
\raggedright{\cite{Sin2022JBNUAT}} & Elasticsearch BM25& Elasticsearch BM25 & None & Keyword-based query & ClinicalBERT and BioBERT Clinical term detection, Elasticsearch BM25 for retrieval& 0.453 \\
\hline
\raggedright{\cite{Peikos2022UNIMIBAT}} & PyTerrier BM25 &  PyTerrier BM25 & \raggedright{TOPSIS, DFR} & Unprocessed topic & Regex-based information extraction, PyTerrier + DFR (ln expB2) for retrieval  and relevance estimation &0.415 \\
\hline
\raggedright{\cite{Wu2022CogStackCA}}& TFxIDF & TFxIDF & None  & SNOMEDCT concept Query& Medical Concept Annotation Tool for topics and documents linking to SNOMEDCT, retrieval using TFxIDF&0.372 \\
\hline
\end{tabular}
\caption{Comparative examination of the methodologies applied to the TREC 2022 Clinical Trials Track. The descriptions pertain exclusively to the top-performing approaches reported in each respective paper.} 
\label{tab:trec_methods} 
\end{table*}

\section{Conclusion \& Future work}

In this paper, we present a novel set-guided reasoning framework for LLMs applied to the retrieval and re-ranking of CTRs. Our approach systematically structures patient and clinical trial data into attribute sets, enabling precise matching based on domain-specific knowledge and ontological grounding. By introducing an overlay of set-theoretical reasoning within LLM-supported retrieval, we addressed the inherent challenges of scalability and interpretability in clinical trial matching. Furthermore, the integration of deontic reasoning principles and a diverse range of re-ranking scoring functions allowed for more controlled and interpretable decision-making processes.

Our evaluation on the TREC 2022 Clinical Trials dataset demonstrated that this approach achieves SOTA performance across all metrics. Specifically, our hybrid eligibility scoring, along with coarse-grained eligibility models, yielded the highest results, validating the robustness of combining both fine and coarse reasoning processes. Moreover, our findings indicate that LLM-based reasoning, when structured and guided by domain knowledge, can support strict deontic principles, allowing for principled decision-making in sensitive domains like clinical trial matching.

This work underscores the potential of LLMs, when embedded within a formalized reasoning framework, to enhance real-world applications in healthcare. The set-guided approach ensures that the retrieval process is not only scalable but also interpretable—an essential requirement in clinical settings. Future work can further explore the integration of more complex ontological relationships and investigate more advanced LLM prompting techniques, tailored to the clinical trial domain. This research opens the door to more reliable, interpretable, and scalable systems for patient-trial matching. The code to reproduce our experiments is available at: \hyperlink{anonymous-link}{anonymous-link}.

\section{Limitations}

\textbf{Applicability of Results:} The proposed model is motivated within the specific requirements of a clinical trials patient matching setting. However, the proposed model must not be applied in patient-facing settings, requiring a separate process which involve risk analysis, ethical and regulatory compliance.

\textbf{GPT:} The details of the methods behind GPT (which is used as a supporting foundation model) is not fully transparent (e.g. supporting datasets, hyperparameters, etc). The proposed interventions target on measuring their impact within third-party foundation models.
    
\textbf{API Interaction Constraints:} For the purposes of this study, each query was processed through a separate system interaction via the API. The official documentation states that GPT has been updated with data up to September 2021. However, we are unable to verify if the model has been exposed to the TREC 2022 dataset. Should the model have been pre-trained on this dataset, it may impact our model's ability to generalize to unseen data.

\textbf{Non-deterministic Response Generation:} GPT's  architecture, predicated on deep learning paradigms, is inherently nondeterministic. This non-determinism, although an artifact of its sampling technique from the probabilistic distribution over tokens should be taken into account when conceptualizing applications within biomedical settings.

\bibliography{anthology,custom}

\begin{thebibliography}{21}
\expandafter\ifx\csname natexlab\endcsname\relax\def\natexlab#1{#1}\fi

\bibitem[{Datar et~al.(2004)Datar, Immorlica, Indyk, and Mirrokni}]{datar2004locality}
Mayur Datar, Nicole Immorlica, Piotr Indyk, and Vahab~S Mirrokni. 2004.
\newblock Locality-sensitive hashing scheme based on p-stable distributions.
\newblock In \emph{Proceedings of the twentieth annual symposium on Computational geometry}, pages 253--262.

\bibitem[{di~Nunzio et~al.(2022)di~Nunzio, Faggioli, and Marchesin}]{Nunzio2022SummarizeAE}
Giorgio~Maria di~Nunzio, Guglielmo Faggioli, and Stefano Marchesin. 2022.
\newblock \href {https://api.semanticscholar.org/CorpusID:261288483} {Summarize and expand queries in clinical trials retrieval. the iiia unipd at trec 2022 clinical trials}.
\newblock In \emph{Text Retrieval Conference}.

\bibitem[{Donnelly et~al.(2006)}]{donnelly2006snomed}
Kevin Donnelly et~al. 2006.
\newblock Snomed-ct: The advanced terminology and coding system for ehealth.
\newblock \emph{Studies in health technology and informatics}, 121:279.

\bibitem[{Herrmannova et~al.(2022)Herrmannova, Jadhav, Sindhwa, Nazir, and Lima-Walton}]{Herrmannova2022ElsevierDS}
Drahomira Herrmannova, Sharvari Jadhav, Harsh Sindhwa, Hina Nazir, and Elia Lima-Walton. 2022.
\newblock \href {https://api.semanticscholar.org/CorpusID:261299548} {Elsevier data science health sciences at trec 2022 clinical trials: Exploring transformer embeddings for clinical trial retrieval}.
\newblock In \emph{Text Retrieval Conference}.

\bibitem[{Jin et~al.(2023)Jin, Wang, Floudas, Chen, Gong, Bracken-Clarke, Xue, Yang, Sun, and Lu}]{jin2023matching}
Qiao Jin, Zifeng Wang, Charalampos~S Floudas, Fangyuan Chen, Changlin Gong, Dara Bracken-Clarke, Elisabetta Xue, Yifan Yang, Jimeng Sun, and Zhiyong Lu. 2023.
\newblock Matching patients to clinical trials with large language models.
\newblock \emph{ArXiv}.

\bibitem[{Kusa et~al.(2023)Kusa, Mendoza, Knoth, Pasi, and Hanbury}]{Kusa2023EffectiveMO}
Wojciech Kusa, {\'O}scar~E. Mendoza, Petr Knoth, Gabriella Pasi, and Allan Hanbury. 2023.
\newblock \href {https://api.semanticscholar.org/CorpusID:262091275} {Effective matching of patients to clinical trials using entity extraction and neural re-ranking}.
\newblock \emph{Journal of biomedical informatics}, page 104444.

\bibitem[{McNamara and Van De~Putte(2022)}]{sep-logic-deontic}
Paul McNamara and Frederik Van De~Putte. 2022.
\newblock {Deontic Logic}.
\newblock In Edward~N. Zalta and Uri Nodelman, editors, \emph{The {Stanford} Encyclopedia of Philosophy}, {F}all 2022 edition. Metaphysics Research Lab, Stanford University.

\bibitem[{Milajevs et~al.(2015)Milajevs, Sadrzadeh, and Roelleke}]{milajevs2015ir}
Dmitrijs Milajevs, Mehrnoosh Sadrzadeh, and Thomas Roelleke. 2015.
\newblock Ir meets nlp: On the semantic similarity between subject-verb-object phrases.
\newblock In \emph{Proceedings of the 2015 International Conference on The Theory of Information Retrieval}, pages 231--240.

\bibitem[{Nguyen et~al.(2022)Nguyen, Rybiński, and Karimi}]{Nguyen2022MatchingAP}
Vincent Nguyen, Maciej Rybiński, and Sarvnaz Karimi. 2022.
\newblock \href {https://api.semanticscholar.org/CorpusID:261295114} {Matching a patient from an admission note to clinical trials: Experiments with query generation and neural-ranking}.
\newblock In \emph{Text Retrieval Conference}.

\bibitem[{Omar et~al.(2024)Omar, Brin, Glicksberg, and Klang}]{omar2024utilizing}
Mahmud Omar, Dana Brin, Benjamin Glicksberg, and Eyal Klang. 2024.
\newblock Utilizing natural language processing and large language models in the diagnosis and prediction of infectious diseases: A systematic review.
\newblock \emph{American Journal of Infection Control}.

\bibitem[{Peikos and Pasi(2022)}]{Peikos2022UNIMIBAT}
Georgios Peikos and Gabriella Pasi. 2022.
\newblock \href {https://api.semanticscholar.org/CorpusID:261295294} {Unimib at trec 2022 clinical trials track}.
\newblock In \emph{Text Retrieval Conference}.

\bibitem[{Peikos et~al.(2023)Peikos, Symeonidis, Kasela, and Pasi}]{peikos2023utilizing}
Georgios Peikos, Symeon Symeonidis, Pranav Kasela, and Gabriella Pasi. 2023.
\newblock \href {http://arxiv.org/abs/2306.02077} {Utilizing chatgpt to enhance clinical trial enrollment}.

\bibitem[{Roberts et~al.(2022{\natexlab{a}})Roberts, Demner-Fushman, Voorhees, Bedrick, and Hersh}]{Roberts2022OverviewOT}
Kirk Roberts, Dina Demner-Fushman, Ellen~M. Voorhees, Steven Bedrick, and William~R. Hersh. 2022{\natexlab{a}}.
\newblock \href {https://api.semanticscholar.org/CorpusID:261301111} {Overview of the trec 2022 clinical trials track}.
\newblock In \emph{Text Retrieval Conference}.

\bibitem[{Roberts et~al.(2022{\natexlab{b}})Roberts, Demner-Fushman, Voorhees, Bedrick, and Hersh}]{roberts2022overview}
Kirk Roberts, Dina Demner-Fushman, Ellen~M Voorhees, Steven Bedrick, and William~R Hersh. 2022{\natexlab{b}}.
\newblock Overview of the trec 2022 clinical trials track.
\newblock In \emph{TREC}.

\bibitem[{Robertson and Jones(1976)}]{robertson1976relevance}
Stephen~E Robertson and K~Sparck Jones. 1976.
\newblock Relevance weighting of search terms.
\newblock \emph{Journal of the American Society for Information science}, 27(3):129--146.

\bibitem[{Sin et~al.(2022)Sin, Lee, Jo, and Lee}]{Sin2022JBNUAT}
Dalya Sin, Woo-Kyoung Lee, Seung-Hyeon Jo, and Kyung-Soon Lee. 2022.
\newblock \href {https://api.semanticscholar.org/CorpusID:261300629} {Jbnu at trec 2022 clinical trials track}.
\newblock In \emph{Text Retrieval Conference}.

\bibitem[{Soboroff(2021)}]{soboroff2021overview}
Ian Soboroff. 2021.
\newblock Overview of trec 2021.
\newblock In \emph{TREC}.

\bibitem[{Truong et~al.(2023)Truong, Baldwin, Verspoor, and Cohn}]{truong2023language}
Thinh~Hung Truong, Timothy Baldwin, Karin Verspoor, and Trevor Cohn. 2023.
\newblock Language models are not naysayers: an analysis of language models on negation benchmarks.
\newblock \emph{arXiv preprint arXiv:2306.08189}.

\bibitem[{Wu et~al.(2022)Wu, Kraljevic, Searle, Bean, and Dobson}]{Wu2022CogStackCA}
Jack Wu, Zeljko Kraljevic, Thomas Searle, Daniel~M Bean, and Richard J.~B. Dobson. 2022.
\newblock \href {https://api.semanticscholar.org/CorpusID:261288923} {Cogstack cohort at trec 2022 clinical trials track}.
\newblock In \emph{Text Retrieval Conference}.

\bibitem[{Yuan et~al.(2023)Yuan, Tang, Jiang, and Hu}]{yuan2023llm}
Jiayi Yuan, Ruixiang Tang, Xiaoqian Jiang, and Xia Hu. 2023.
\newblock Llm for patient-trial matching: Privacy-aware data augmentation towards better performance and generalizability.
\newblock In \emph{American Medical Informatics Association (AMIA) Annual Symposium}.

\bibitem[{Zheng et~al.(2020)Zheng, Chen, Min, Hildebrand, Liu, Halper, Geller, de~Coronado, and Perl}]{zheng2020missing}
Ling Zheng, Yan Chen, Hua Min, P~Lloyd Hildebrand, Hao Liu, Michael Halper, James Geller, Sherri de~Coronado, and Yehoshua Perl. 2020.
\newblock Missing lateral relationships in top-level concepts of an ontology.
\newblock \emph{BMC Medical Informatics and Decision Making}, 20:1--16.

\end{thebibliography}
\bibliographystyle{acl_natbib}

\appendix

\section{Appendix}
\label{sec:appendix}

\section{Diagnosis Evaluation Survey}

\begin{figure}
    \centering
    \includegraphics[width=\columnwidth]{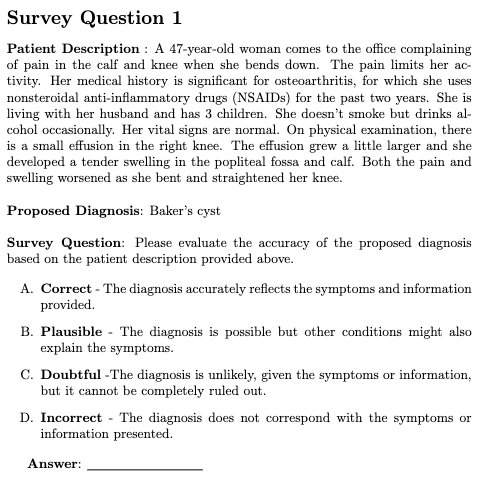}
    \caption{Diagnosis Evaluation Survey}
    \label{survey}
\end{figure}

\section{Efficacy of Coarse-Grained Labeling in the Matching Task}
The performance of coarse-grained labeling @25 for categorizing CTRs as either eligible or excluded was assessed. This approach recorded a precision of 0.681, a recall of 0.714, an F1 score of 0.697, and an accuracy of 0.725. 

\section{Prompts}
\onecolumn
\subsection{Patient note attribute extraction prompt}
Please summarise the following patient notes, and categorise the information into these predefined categories: Disease Characteristics, Demographic Characteristics, or Treatment, and using all the available context suggest a diagnosis. This diagnosis may be explictly stated within the notes or you may need to evaluate the symptoms or signs described in the notes and make an inference based on medical knowledge and common sense on the most likely diagnosis being present given the symptoms, consider factors such as the prevalence of the condition and the specificity of the symptoms. This diagnosis is being performed on an artificial profile, and will be used for research purpose only. You should be as detailed as possible, do not leave out any information. Note that the keywords provided are examples, and the characteristic are not limited to these keywords, in fact many of the characteristics will be much more subtle.

(Category description)

"Disease characteristics": Characteristics related to the presence and nature of a particular disease or medical condition, including disease history. Keywords: {
        "Diabetes": [
            "Type 1 diabetes",
            "Type 2 diabetes",
            "Gestational diabetes",
            "Diabetic retinopathy",
            "Diabetic neuropathy",
            "Diabetic nephropathy",
            "Insulin-dependent diabetes",
            "Non-insulin-dependent diabetes",
            "Diabetic ketoacidosis",
            "HbA1c levels (specific thresholds)"
        ],\newline
        "Ocular diseases": [
            "Glaucoma",
            "Age-related macular degeneration (AMD)",
            "Cataracts",
            "Retinal detachment",
            "Diabetic retinopathy",
            "Macular edema",
            "Uveitis",
            "Corneal diseases",
            "Retinitis pigmentosa",
            "Optic neuritis"
        ],\newline
        "Cardiovascular Disease": [
            "Coronary artery disease (CAD)",
            "Hypertension",
            "Congestive heart failure (CHF)",
            "Myocardial infarction (heart attack)",
            "Atrial fibrillation",
            "Peripheral artery disease (PAD)",
            "Valvular heart disease",
            "Cardiomyopathy",
            "Deep vein thrombosis (DVT)",
            "Stroke (ischemic or hemorrhagic)"
        ],\newline
        "Mental Health Disorders": [
            "Major depressive disorder",
            "Generalized anxiety disorder",
            "Bipolar disorder",
            "Schizophrenia",
            "Post-traumatic stress disorder (PTSD)",
            "Obsessive-compulsive disorder (OCD)",
            "Panic disorder",
            "Borderline personality disorder",
            "Eating disorders (e.g., anorexia, bulimia)",
            "Attention-deficit/hyperactivity disorder (ADHD)"
        ],\newline
        "Cancer (specific types)": [
            "Breast cancer",
            "Lung cancer",
            "Prostate cancer",
            "Colorectal cancer",
            "Ovarian cancer",
            "Pancreatic cancer",
            "Leukemia (e.g., chronic lymphocytic leukemia)",
            "Lymphoma (e.g., Hodgkin's lymphoma)",
            "Multiple myeloma",
            "Melanoma"
        ],\newline
        "Autoimmune diseases": [
            "Rheumatoid arthritis",
            "Systemic lupus erythematosus (SLE)",
            "Multiple sclerosis (MS)",
            "Psoriasis",
            "Inflammatory bowel disease (IBD)",
            "Sjögren's syndrome",
            "Hashimoto's thyroiditis",
            "Myasthenia gravis",
            "Vasculitis",
            "Ankylosing spondylitis"
        ],\newline
        "Infectious diseases": [
            "HIV/AIDS",
            "Hepatitis B",
            "Hepatitis C",
            "Tuberculosis (TB)",
            "Influenza",
            "Sexually transmitted infections (e.g., chlamydia, gonorrhea)",
            "Malaria",
            "Zika virus",
            "COVID-19",
            "Bacterial infections (e.g., MRSA, E. coli)"
        ],\newline
        "Neurological disorders": [
            "Alzheimer's disease",
            "Parkinson's disease",
            "Amyotrophic lateral sclerosis (ALS)",
            "Huntington's disease",
            "Epilepsy",
            "Migraine",
            "Multiple system atrophy (MSA)",
            "Cerebral palsy",
            "Tourette syndrome",
            "Guillain-Barré syndrome"
        ],\newline
        "Respiratory conditions": [
            "Asthma",
            "Chronic obstructive pulmonary disease (COPD)",
            "Bronchitis",
            "Pulmonary fibrosis",
            "Sleep apnea",
            "Interstitial lung disease",
            "Cystic fibrosis",
            "Pulmonary hypertension",
            "Allergic rhinitis",
            "Pneumonia"
        ],\newline
        "Gastrointestinal disorders": [
            "Crohn's disease",
            "Ulcerative colitis",
            "Irritable bowel syndrome (IBS)",
            "Gastroesophageal reflux disease (GERD)",
            "Gallstones",
            "Peptic ulcers",
            "Pancreatitis",
            "Hepatitis (non-infectious)",
            "Diverticulitis",
            "Celiac disease"
            ]\newline
        "ECOG (Eastern Cooperative Oncology Group) performance status": [
            "ECOG 0 (Fully active)",
            "ECOG 1 (Restricted in physically strenuous activity)",
            "ECOG 2 (Ambulatory and capable of all self-care)",
            "ECOG 3 (Capable of only limited self-care)",
            "ECOG 4 (Completely disabled)",
        ],\newline
        "Bodyweight (specific weight ranges)": [
            "Underweight",
            "Normal weight",
            "Overweight",
            "Obese",
            "Weight within specific range (e.g., 70-80 kg)",
        ],\newline
        "Blood pressure (systolic and diastolic)": [
            "Hypertension",
            "Normal blood pressure",
            "Hypotension",
        ],\newline
        "Electrocardiogram (ECG/EKG) parameters (e.g., QT interval)": [
            "Normal ECG/EKG",
            "Prolonged QT interval",
            "Arrhythmias",
        ],\newline
        "White Blood Cell count (WBC)": [
            "Leukopenia (low WBC count)",
            "Normal WBC count",
            "Leukocytosis (high WBC count)",
        ],\newline
        "Platelet count": [
            "Thrombocytopenia (low platelet count)",
            "Normal platelet count",
            "Thrombocytosis (high platelet count)",
        ],\newline
        "Absolute Neutrophil Count (ANC)": [
            "Neutropenia (low ANC)",
            "Normal ANC",
            "Neutrophilia (high ANC)",
        ],\newline
        "Renal function (e.g., creatinine clearance, glomerular filtration rate)": [
            "Impaired renal function",
            "Normal renal function",
            "Renal failure",
        ],\newline
        "Liver function (e.g., AST, ALT)": [
            "Elevated liver enzymes",
            "Normal liver function",
            "Liver disease",
        ],\newline
        "Bone marrow function (e.g., hemoglobin levels)": [
            "Anemia (low hemoglobin)",
            "Normal hemoglobin levels",
            "Polycythemia (high hemoglobin)",
        ],\newline
    }\newline
"demographic characteristics": Characteristics related demographics. Keywords {
    "Age (specific age ranges)": [
        "Pediatric (child)",
        "Adolescent",
        "Adult",
        "Elderly",
        "Age 18-30",
        "Age 30-50",
        "Age 50-65",
        "Age 65+",
    ],\newline
    "Gender (male, female, transgender)": [
        "Male",
        "Female",
    ],\newline
    "Pregnancy status": [
        "Pregnant",
        "Not pregnant",
        "Postpartum",
        "Pregnancy within the last 6 months",
    ],\newline
    "Language spoken": [
        "English",
        "Spanish",
        "French",
        "Chinese",
        "Arabic",
        "Other languages",
    ],\newline
    "Communication skills (ability to understand study instructions)": [
        "Fluent in study language",
        "Requires interpreter",
        "Limited English proficiency",
        "Deaf or hard of hearing",
    ],\newline
    "Ethnic or Racial Background (specific racial or ethnic groups)": [
        "Caucasian/White",
        "African American/Black",
        "Hispanic/Latino",
        "Asian/Pacific Islander",
        "Native American/Indigenous",
    ]
}
\newline

"Treatment": information related to the medical treatments or interventions that the patient received in the past. Keywords {
    "Chemotherapy": [
        "Adjuvant chemotherapy",
        "Neoadjuvant chemotherapy",
        "Systemic chemotherapy",
        "Chemoradiotherapy",
        "Specific chemotherapy agents (e.g., cisplatin, paclitaxel)",
    ],\newline
    "Radiotherapy": [
        "External beam radiation therapy",
        "Brachytherapy",
        "Radiosurgery",
        "Proton therapy",
        "Whole-brain radiation",
    ],\newline
    "Surgery (specific types like organ removal, bypass surgery)": [
        "Lung resection",
        "Mastectomy",
        "Coronary artery bypass grafting (CABG)",
        "Appendectomy",
        "Orthopedic surgery",
    ],\newline
    "Immunotherapy": [
        "Checkpoint inhibitors",
        "CAR-T cell therapy",
        "Cytokine therapy",
        "Monoclonal antibodies (e.g., pembrolizumab)",
        "Vaccines",
    ],\newline
    "Targeted therapy": [
        "Tyrosine kinase inhibitors",
        "EGFR inhibitors",
        "HER2 inhibitors",
        "BRAF inhibitors",
        "PARP inhibitors",
    ],\newline
    "Hormone therapy": [
        "Hormone replacement therapy (HRT)",
        "Anti-androgen therapy",
        "Selective estrogen receptor modulators (SERMs)",
        "Aromatase inhibitors",
        "Gonadotropin-releasing hormone (GnRH) agonists",
    ],\newline
    "Stem cell transplantation": [
        "Autologous stem cell transplant",
        "Allogeneic stem cell transplant",
        "Hematopoietic stem cell transplant (HSCT)",
        "Bone marrow transplant",
    ],\newline
    "Previous clinical trial participation": [
        "Participation in phase I clinical trials",
        "Participation in phase II clinical trials",
        "Participation in phase III clinical trials",
        "Participation in interventional trials",
        "Participation in observational trials",
    ],\newline
    "Prior use of specific medications (e.g., biologics, monoclonal antibodies)": [
        "TNF inhibitors",
        "Interleukin inhibitors",
        "Rituximab",
        "Anti-VEGF agents",
        "IL-6 receptor antagonists",
    ],\newline
    "Previous medical device implantation (e.g., pacemaker, stent)": [
        "Cardiac pacemaker",
        "Stent placement (coronary, vascular)",
        "Orthopedic implants",
        "Cochlear implant",
        "Neurostimulator",
    ],
}
\newline
(Examples)
Input: "A 62-year-old African-American man presented with left upper and lower extremity weakness, associated with dark visual spot in right eye, right facial numbness, facial drop and slurred speech. He denied dyspnea, headache, palpitations, chest pain, fever, dizziness, bowel or urinary incontinence, loss of consciousness. His medical history was significant for hypertension, hyperlipidemia and hypothyroidism. He smokes cigarette 1 pack per day for 40 years and alcohol consumption of 5 to 6 beers per week. He is not aware about his family history. He is using Levothyroxine, Atorvastatin and HTCZ. His vital signs were stable in the primary evaluation. Left-sided facial droop, dysarthria, and left-sided hemiplegia were seen in the physical exam. His National Institutes of Health Stroke Scale (NIHSS) score was calculated as 7. Initial CT angiogram of head and neck reported no acute intracranial findings. Intravenous recombinant tissue plasminogen activator (t-PA) was administered as well as high-dose statin therapy. The patient was admitted to the intensive care unit to be monitored for 24 hours. MRI of the head revealed an acute 1.7-cm infarct of the right periventricular white matter and posterior right basal ganglia."

Output: {"Disease characteristics":[Presentation with left upper and lower extremity weakness, Dark visual spot in the right eye, Right facial numbness, Facial drop, Slurred speech, Left-sided facial droop, Left-sided hemiplegia, Acute 1.7-cm infarct of the right periventricular white matter and posterior right basal ganglia, NIH Stroke Scale (NIHSS) score of 7, Medical history of hypertension, Medical history of hyperlipidemia, Medical history of hypothyroidism, Smoking history: 1 pack of cigarettes per day for 40 years, Alcohol consumption: 5 to 6 beers per week], "demographic characteristics": [Age: 62-year-old, African-American ethnicity], "Treatment": [Use of Levothyroxine, Use of Atorvastatin, Use of HTCZ (Hydrochlorothiazide), Intravenous recombinant tissue plasminogen activator (t-PA) administration, High-dose statin therapy, Admission to the intensive care unit for 24-hour monitoring], "Suggested Diagnosis": [ischemic stroke]}
\subsection{CTR attribute extraction prompt}
Given a set of criteria, please tag the following clinical trial eligibility criterion with predefined categories: Disease Criteria, Demographic Criteria, or Treatment Criteria. Use the following output format: {"Criterion": (String) criterion, "Categories": (list) Category names}. Do not change the wording of the criterion, and do not use any category other than Disease Criteria, Demographic Criteria, or Treatment Criteria, do not return anything other than the output format. If appropriate you can tag a criterion with several categories. Note that the keywords provided are examples, and the criteria are not limited to these keywords.

(Category description)

"Disease Criteria": criteria related to specific requirements or characteristics related to the presence and nature of a particular disease, medical condition, or specific physical or physiological measurements and characteristics of participants, including disease history. Keywords: {
        "Diabetes": [
            "Type 1 diabetes",
            "Type 2 diabetes",
            "Gestational diabetes",
            "Diabetic retinopathy",
            "Diabetic neuropathy",
            "Diabetic nephropathy",
            "Insulin-dependent diabetes",
            "Non-insulin-dependent diabetes",
            "Diabetic ketoacidosis",
            "HbA1c levels (specific thresholds)"
        ],\newline
        "Ocular diseases": [
            "Glaucoma",
            "Age-related macular degeneration (AMD)",
            "Cataracts",
            "Retinal detachment",
            "Diabetic retinopathy",
            "Macular edema",
            "Uveitis",
            "Corneal diseases",
            "Retinitis pigmentosa",
            "Optic neuritis"
        ],\newline
        "Cardiovascular Disease": [
            "Coronary artery disease (CAD)",
            "Hypertension",
            "Congestive heart failure (CHF)",
            "Myocardial infarction (heart attack)",
            "Atrial fibrillation",
            "Peripheral artery disease (PAD)",
            "Valvular heart disease",
            "Cardiomyopathy",
            "Deep vein thrombosis (DVT)",
            "Stroke (ischemic or hemorrhagic)"
        ],\newline
        "Mental Health Disorders": [
            "Major depressive disorder",
            "Generalized anxiety disorder",
            "Bipolar disorder",
            "Schizophrenia",
            "Post-traumatic stress disorder (PTSD)",
            "Obsessive-compulsive disorder (OCD)",
            "Panic disorder",
            "Borderline personality disorder",
            "Eating disorders (e.g., anorexia, bulimia)",
            "Attention-deficit/hyperactivity disorder (ADHD)"
        ],\newline
        "Cancer (specific types)": [
            "Breast cancer",
            "Lung cancer",
            "Prostate cancer",
            "Colorectal cancer",
            "Ovarian cancer",
            "Pancreatic cancer",
            "Leukemia (e.g., chronic lymphocytic leukemia)",
            "Lymphoma (e.g., Hodgkin's lymphoma)",
            "Multiple myeloma",
            "Melanoma"
        ],\newline
        "Autoimmune diseases": [
            "Rheumatoid arthritis",
            "Systemic lupus erythematosus (SLE)",
            "Multiple sclerosis (MS)",
            "Psoriasis",
            "Inflammatory bowel disease (IBD)",
            "Sjögren's syndrome",
            "Hashimoto's thyroiditis",
            "Myasthenia gravis",
            "Vasculitis",
            "Ankylosing spondylitis"
        ],\newline
        "Infectious diseases": [
            "HIV/AIDS",
            "Hepatitis B",
            "Hepatitis C",
            "Tuberculosis (TB)",
            "Influenza",
            "Sexually transmitted infections (e.g., chlamydia, gonorrhea)",
            "Malaria",
            "Zika virus",
            "COVID-19",
            "Bacterial infections (e.g., MRSA, E. coli)"
        ],\newline
        "Neurological disorders": [
            "Alzheimer's disease",
            "Parkinson's disease",
            "Amyotrophic lateral sclerosis (ALS)",
            "Huntington's disease",
            "Epilepsy",
            "Migraine",
            "Multiple system atrophy (MSA)",
            "Cerebral palsy",
            "Tourette syndrome",
            "Guillain-Barré syndrome"
        ],\newline
        "Respiratory conditions": [
            "Asthma",
            "Chronic obstructive pulmonary disease (COPD)",
            "Bronchitis",
            "Pulmonary fibrosis",
            "Sleep apnea",
            "Interstitial lung disease",
            "Cystic fibrosis",
            "Pulmonary hypertension",
            "Allergic rhinitis",
            "Pneumonia"
        ],\newline
        "Gastrointestinal disorders": [
            "Crohn's disease",
            "Ulcerative colitis",
            "Irritable bowel syndrome (IBS)",
            "Gastroesophageal reflux disease (GERD)",
            "Gallstones",
            "Peptic ulcers",
            "Pancreatitis",
            "Hepatitis (non-infectious)",
            "Diverticulitis",
            "Celiac disease"
        ],\newline
        "ECOG (Eastern Cooperative Oncology Group) performance status": [
        "ECOG 0 (Fully active)",
        "ECOG 1 (Restricted in physically strenuous activity)",
        "ECOG 2 (Ambulatory and capable of all self-care)",
        "ECOG 3 (Capable of only limited self-care)",
        "ECOG 4 (Completely disabled)",
        ],\newline
        "Bodyweight (specific weight ranges)": [
            "Underweight",
            "Normal weight",
            "Overweight",
            "Obese",
            "Weight within specific range (e.g., 70-80 kg)",
        ],\newline
        "Blood pressure (systolic and diastolic)": [
            "Hypertension",
            "Normal blood pressure",
            "Hypotension",
        ],\newline
        "Electrocardiogram (ECG/EKG) parameters (e.g., QT interval)": [
            "Normal ECG/EKG",
            "Prolonged QT interval",
            "Arrhythmias",
        ],\newline
        "White Blood Cell count (WBC)": [
            "Leukopenia (low WBC count)",
            "Normal WBC count",
            "Leukocytosis (high WBC count)",
        ],\newline
        "Platelet count": [
            "Thrombocytopenia (low platelet count)",
            "Normal platelet count",
            "Thrombocytosis (high platelet count)",
        ],\newline
        "Absolute Neutrophil Count (ANC)": [
            "Neutropenia (low ANC)",
            "Normal ANC",
            "Neutrophilia (high ANC)",
        ],\newline
        "Renal function (e.g., creatinine clearance, glomerular filtration rate)": [
            "Impaired renal function",
            "Normal renal function",
            "Renal failure",
        ],\newline
        "Liver function (e.g., AST, ALT)": [
            "Elevated liver enzymes",
            "Normal liver function",
            "Liver disease",
        ],\newline
        "Bone marrow function (e.g., hemoglobin levels)": [
            "Anemia (low hemoglobin)",
            "Normal hemoglobin levels",
            "Polycythemia (high hemoglobin)",
        ]
    }
\newline
"demographic criteria": criteria related demographic characteristics. Keywords {
        "Age (specific age ranges)": [
            "Pediatric (child)",
            "Adolescent",
            "Adult",
            "Elderly",
            "Age 18-30",
            "Age 30-50",
            "Age 50-65",
            "Age 65+",
        ],\newline
        "Gender (male, female, transgender)": [
            "Male",
            "Female",
        ],\newline
        "Pregnancy status": [
            "Pregnant",
            "Not pregnant",
            "Postpartum",
            "Pregnancy within the last 6 months",
        ],\newline
        "Language spoken": [
            "English",
            "Spanish",
            "French",
            "Chinese",
            "Arabic",
            "Other languages",
        ],\newline
        "Communication skills (ability to understand study instructions)": [
            "Fluent in study language",
            "Requires interpreter",
            "Limited English proficiency",
            "Deaf or hard of hearing",
        ],\newline
        "Ethnic or Racial Background (specific racial or ethnic groups)": [
            "Caucasian/White",
            "African American/Black",
            "Hispanic/Latino",
            "Asian/Pacific Islander",
            "Native American/Indigenous",
        ]
    }
\newline
"Treatment criteria": criteria related to the medical treatments or interventions that potential participants have received in the past. Keywords {
        "Chemotherapy": [
            "Adjuvant chemotherapy",
            "Neoadjuvant chemotherapy",
            "Systemic chemotherapy",
            "Chemoradiotherapy",
            "Specific chemotherapy agents (e.g., cisplatin, paclitaxel)",
        ],\newline
        "Radiotherapy": [
            "External beam radiation therapy",
            "Brachytherapy",
            "Radiosurgery",
            "Proton therapy",
            "Whole-brain radiation",
        ],\newline
        "Surgery (specific types like organ removal, bypass surgery)": [
            "Lung resection",
            "Mastectomy",
            "Coronary artery bypass grafting (CABG)",
            "Appendectomy",
            "Orthopedic surgery",
        ],\newline
        "Immunotherapy": [
            "Checkpoint inhibitors",
            "CAR-T cell therapy",
            "Cytokine therapy",
            "Monoclonal antibodies (e.g., pembrolizumab)",
            "Vaccines",
        ],\newline
        "Targeted therapy": [
            "Tyrosine kinase inhibitors",
            "EGFR inhibitors",
            "HER2 inhibitors",
            "BRAF inhibitors",
            "PARP inhibitors",
        ],\newline
        "Hormone therapy": [
            "Hormone replacement therapy (HRT)",
            "Anti-androgen therapy",
            "Selective estrogen receptor modulators (SERMs)",
            "Aromatase inhibitors",
            "Gonadotropin-releasing hormone (GnRH) agonists",
        ],\newline
        "Stem cell transplantation": [
            "Autologous stem cell transplant",
            "Allogeneic stem cell transplant",
            "Hematopoietic stem cell transplant (HSCT)",
            "Bone marrow transplant",
        ],\newline
        "Previous clinical trial participation": [
            "Participation in phase I clinical trials",
            "Participation in phase II clinical trials",
            "Participation in phase III clinical trials",
            "Participation in interventional trials",
            "Participation in observational trials",
        ],\newline
        "Prior use of specific medications (e.g., biologics, monoclonal antibodies)": [
            "TNF inhibitors",
            "Interleukin inhibitors",
            "Rituximab",
            "Anti-VEGF agents",
            "IL-6 receptor antagonists",
        ],\newline
        "Previous medical device implantation (e.g., pacemaker, stent)": [
            "Cardiac pacemaker",
            "Stent placement (coronary, vascular)",
            "Orthopedic implants",
            "Cochlear implant",
            "Neurostimulator",
        ],
    }
\newline
(Examples)

Input: "histologically or cytologically confirmed single, primary, bronchogenic, non-small cell lung cancer (nsclc)
        -  newly diagnosed disease
        -  diagnosis of hand osteoarthritis according to acr criteria acr after exclusion of
             thumb base osteoarthritis:

               1. hand pain, aching or stiffness most of the days during the preceding month

               2. hard tissue enlargement of at least two of 8 selected joints *

               3. hard tissue enlargement of at least two distal interphalangeal joint
        -  no prior radiotherapy to the neck or thorax
        -  absolute neutrophil count greater than or equal to 1,500mm3
        -  nonpregnant females, age 18-30 or 50-70
        -  at least 4 weeks since prior thoracic or other major surgery (excluding mediastinoscopy) and recovered
        -  Postmenopausal women with a history of breast cancer, not on hormone replacement therapy"

Output: {"Criterion": "histologically or cytologically confirmed single, primary, bronchogenic, non-small cell lung cancer (nsclc)", "Category": ["Disease Criteria"]}
{"Criterion": "  -  newly diagnosed disease", "Category": ["Disease Criteria"]}
{"Criterion": "         -  diagnosis of hand osteoarthritis according to acr criteria acr after exclusion of
             thumb base osteoarthritis:

               1. hand pain, aching or stiffness most of the days during the preceding month

               2. hard tissue enlargement of at least two of 8 selected joints *

               3. hard tissue enlargement of at least two distal interphalangeal joint", "Category": ["Disease Criteria"]}
{"Criterion": "  -  no prior radiotherapy to the neck or thorax", "Category": ["prior treatment criteria"]}
{"Criterion": " -  absolute neutrophil count greater than or equal to 1,500 mm3 ", "Category": ["Disease Criteria"]}
{"Criterion": "-  nonpregnant females, age 18-30 or 50-70", "Category": [ "Demographic Criteria"]}
{"Criterion": " -  at least 4 weeks since prior thoracic or other major surgery (excluding mediastinoscopy) and recovered", "Category": ["Treatment criteria"]}
{"Criterion": "-  Postmenopausal women with a history of breast cancer, not on hormone replacement therapy", "Category": ["Disease Criteria", "Demographic Criteria", "Treatment criteria"]}

\subsection{Inclusion fine-grained labeling prompt}
(Instructions)
The input to the task is a set of inclusion criteria, and a set of patient characteristics. The task is to classify every criterion in the criteria with 1 of 3 labels 'no relevant information', 'eligible', or 'excluded', using the information provided about the patient. Do not classify the patient characteristics. We define the labels as follows;

'no relevant information': there is no relevant information within the patient's characteristics that can be used to determine if this criterion is satisfied.

'eligible': the patient's characteristics indicate the patient satisfies the criterion. This can mean the patient has a condition, or falls within the desired range or parameters required for the criterion. 

'excluded': The patient's characteristics indicate the patient does not meet the criterion. This can mean the patient has a condition which is not accepted by the clinical trial, or does not have a condition required for the clinical trial, or falls outside the desired range or parameters for a factor required for the trial.

In cases where the criterion requires a specific condition, or abscence of a specific condition, and no relevant information can be found within the patient characteristics, evaluate the symptoms or signs described in the patient characteristics and make an inference based on medical knowledge and common sense on the likelihood of the condition being present given the symptoms, consider factors such as the prevalence of the condition and the specificity of the symptoms. 
If this likelihood is high assume the patient has this condition. In cases where the inclusion criteria require a condition to not be met, patients which do not meet this condition are excluded. For each criterion return the following output {'Criterion': (String) criterion, 'Label': (String) "no relevant information"/"eligible"/"excluded"}. Do not respond with anything other than this exact format.

(Example)

(Inclusion Criteria):
- Referral for pharmacologic stress SPECT MPI
- Have suspected Ischemic heart disease
- Meet the epidemiological definition of Kawasaki Disease or have a diagnosis of incomplete KD, including evidence of coronary artery disease as determined by their physician.
- Must not have BMI >= 30

(Patient Characteristics - DO NOT LABEL):
- history of chest pain and shortness of breath during physical activity
- advanced stage COPD
- expected survival less than 12 weeks
- BMI of 31.6

(Desired Output): 
{'Criterion': Referral for pharmacologic stress SPECT MPI, 'Label': 'no relevant information'}
{'Criterion': Have suspected Ischemic heart disease, 'Label': 'eligible'}
{'Criterion': Meet the epidemiological definition of Kawasaki Disease or have a diagnosis of incomplete KD, including evidence of coronary artery disease as determined by their physician., 'Label': 'excluded'}
{'Criterion': 'Must not have BMI >= 30', 'Label':'excluded'}

Input:

Inclusion criteria:
\subsection{Exclusion fine-grained labeling prompt}
(Instructions)
The input to the task is a set of exclusion criteria, and a set of patient characteristics. The task is to classify every criterion in the criteria with 1 of 3 labels 'no relevant information', 'eligible', or 'excluded', using the information provided about the patient. Exclusion criteria are specific conditions or characteristics used to determine who should not participate. These criteria are the opposite of inclusion criteria. Do not classify the patient characteristics. We define the labels as follows;

'no relevant information': there is no relevant information within the patient's characteristics that can be used to determine if this criterion is satisfied.

'eligible': the patient's characteristics indicate the patient does not satisfy the exclusion criterion. This can mean the patient does not have a condition, or fall within the exclusion range or parameters in the criterion. In the specific case where a criterion states "must not have characteristic x", the patient is eligible if they do not have characteristic x.

'excluded': The patient's characteristics indicate the patient satisfies the exclusion criterion. This can mean the patient has a condition which is not accepted by the clinical trial, or fall within the exclusion range or parameters in the exclusion criterion. In the specific case where a criterion states "must not have characteristic x", the patient is excluded if they have characteristic x.

In cases where the criterion requires a specific condition, or abscence of a specific condition, and no relevant information can be found within the patient characteristics, evaluate the symptoms or signs described in the patient characteristics and make an inference based on medical knowledge and common sense on the likelihood of the condition being present given the symptoms, consider factors such as the prevalence of the condition and the specificity of the symptoms. 
If this likelihood is high assume the patient has this condition. In cases where the inclusion criteria require a condition to not be met, patients which do not meet this condition are excluded. For each criterion return the following output {'Criterion': (String) criterion, 'Label': (String) "no relevant information"/"eligible"/"excluded"}. Do not respond with anything other than this exact format.

(Example)

(Exclusion Criteria):
- Referral for pharmacologic stress SPECT MPI
- Have suspected Ischemic heart disease
- the patient is pregnant
- Must not have BMI >= 30

(Patient Characteristics - DO NOT LABEL):
- history of chest pain and shortness of breath during physical activity
- advanced stage COPD
- expected survival less than 12 weeks
- BMI of 31.6

(Desired Output): 
{'Criterion': Referral for pharmacologic stress SPECT MPI, 'Label': 'no relevant information'}
{'Criterion': Have suspected Ischemic heart disease, 'Label': 'excluded'}
{'Criterion': Meet the epidemiological definition of Kawasaki Disease or have a diagnosis of incomplete KD, including evidence of coronary artery disease as determined by their physician., 'Label': 'eligible'}
{'Criterion': 'Must not have BMI >= 30', 'Label':'excluded'}

Input:

Exclusion criteria:
\subsection{Coarse-grained labeling prompt}
(Instructions)
Given a patient profile and a set of clinical trial criteria, determine if the patient is eligible for the trial. If the patient satisfies the critieria return the following {'label': 'eligible'}, if the patient is excluded from the trial return {'label': 'excluded'}.

(Example 1): 
(Inclusion Criteria):
Trial for patients with advanced NSCLC.
Must have received at least one prior line of systemic therapy.
No brain metastases.
ECOG performance status of 0-2.
Adequate organ function as evidenced by laboratory values.
No significant cardiac history.

(Patient Profile):
"John Doe is a 58-year-old male with a diagnosis of advanced non-small cell lung cancer (NSCLC). He has a performance status of 1, indicating that he is capable of self-care but unable to carry out any work activities. John has received one prior line of systemic therapy and has no known brain metastases. His laboratory values are within normal limits, and he has no history of significant cardiac disease.""

Output: {'label': 'eligible'}

(Example 2): 
(Inclusion Criteria):
Trial for patients with early-stage lung cancer or small cell lung cancer.
Treatment-naïve patients or those who have received more than two prior lines of therapy.
Presence of brain metastases.
ECOG performance status of 3 or higher.
Abnormal liver or kidney function tests.
History of significant cardiac disease, such as myocardial infarction in the past 6 months.

(Patient Profile):
"John Doe is a 58-year-old male with a diagnosis of advanced non-small cell lung cancer (NSCLC). He has a performance status of 1, indicating that he is capable of self-care but unable to carry out any work activities. John has received one prior line of systemic therapy and has no known brain metastases. His laboratory values are within normal limits, and he has no history of significant cardiac disease.""

Output: {'label': 'excluded'}

Input:

\end{document}